\pdfoutput=1
\RequirePackage[table]{xcolor}
\documentclass{article}
\usepackage{wrapfig}
\usepackage[preprint]{corl_2026}

\usepackage{multicol}
\usepackage{float}
\usepackage{graphicx}
\usepackage{amsmath}
\usepackage{amssymb}
\usepackage{amsfonts}
\usepackage{multirow}
\usepackage{siunitx}
\usepackage{makecell}
\usepackage{subcaption}
\usepackage{booktabs}
\usepackage{array}
\usepackage{tabularx}
\usepackage{pifont}
\usepackage{needspace}
\usepackage{algorithm}
\usepackage{algorithmic}
\usepackage{enumitem}

\definecolor{myblue}{rgb}{0.88,0.98,1}
\newcommand{\subhead}[1]{\textcolor[gray]{0.25}{\emph{#1}}}

\newcommand{\tablestyle}{%
    \fontsize{8}{10}\selectfont
    \renewcommand{\arraystretch}{1.05}%
    \renewcommand{\cellset}{\renewcommand{\arraystretch}{1.05}}%
    \centering
    \captionsetup{skip=3pt}%
    \captionsetup*[subtable]{skip=2pt}%
}
\newsavebox{\metricstablebox}
\newcommand{\metricstable}[3]{%
    \begingroup
    \setlength{\tabcolsep}{0pt}%
    \sbox{\metricstablebox}{\begin{tabular}{#1}#3\end{tabular}}%
    \setlength{\tabcolsep}{\dimexpr(\linewidth-\wd\metricstablebox)/#2/2\relax}%
    \begin{tabular}{#1}#3\end{tabular}%
    \endgroup
}

\hypersetup{pdftitle={RoboKit: Accelerating Parallel Differentiable Optimization for Robot Learning},pdfauthor={Jianglong Ye, Xuanbin Peng, Conor Mc Gartoll, Changwei Jing, Lai Wei, Jiali Chen, Xueyan Zou, Xiaolong Wang},pdfsubject={}}
\begin{document}

\title{RoboKit: Accelerating Parallel Differentiable Optimization for Robot Learning}

\author{
\normalfont Jianglong Ye$^{*\,\dagger}$, Xuanbin Peng$^{*}$, Conor Mc Gartoll$^{*}$, Changwei Jing$^{*}$\\
\normalfont Lai Wei, Jiali Chen, Xueyan Zou, Xiaolong Wang\\[2pt]
UC San Diego
}

\maketitle
{\let\thefootnote\relax\footnotetext{$^*$Equal contribution.\quad $^\dagger$Project lead.}}

\begin{nolinenumbers}
    \nointerlineskip
    \vbox to 0pt{%
            \vskip -0.28in
            \hbox to \textwidth{\hfil
                \raisebox{-0.25\height}{\includegraphics[height=1.35em]{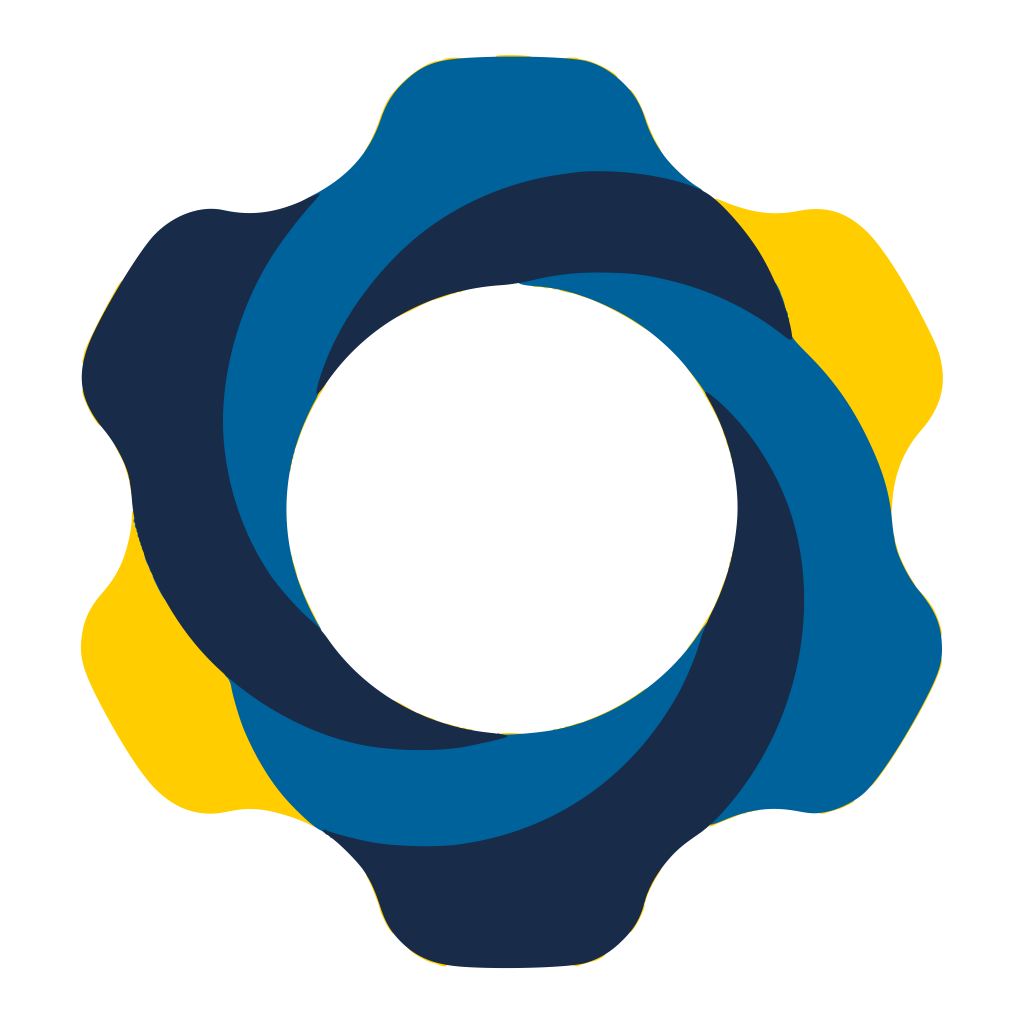}}%
                \hspace{0.4em}\url{https://openrobokit.com/robokit}\hfil}%
            \vss
        }
\end{nolinenumbers}

\vspace{1em}

\begin{figure}[H]
    \centering
    \vspace{-2em}
    \includegraphics[width=\linewidth]{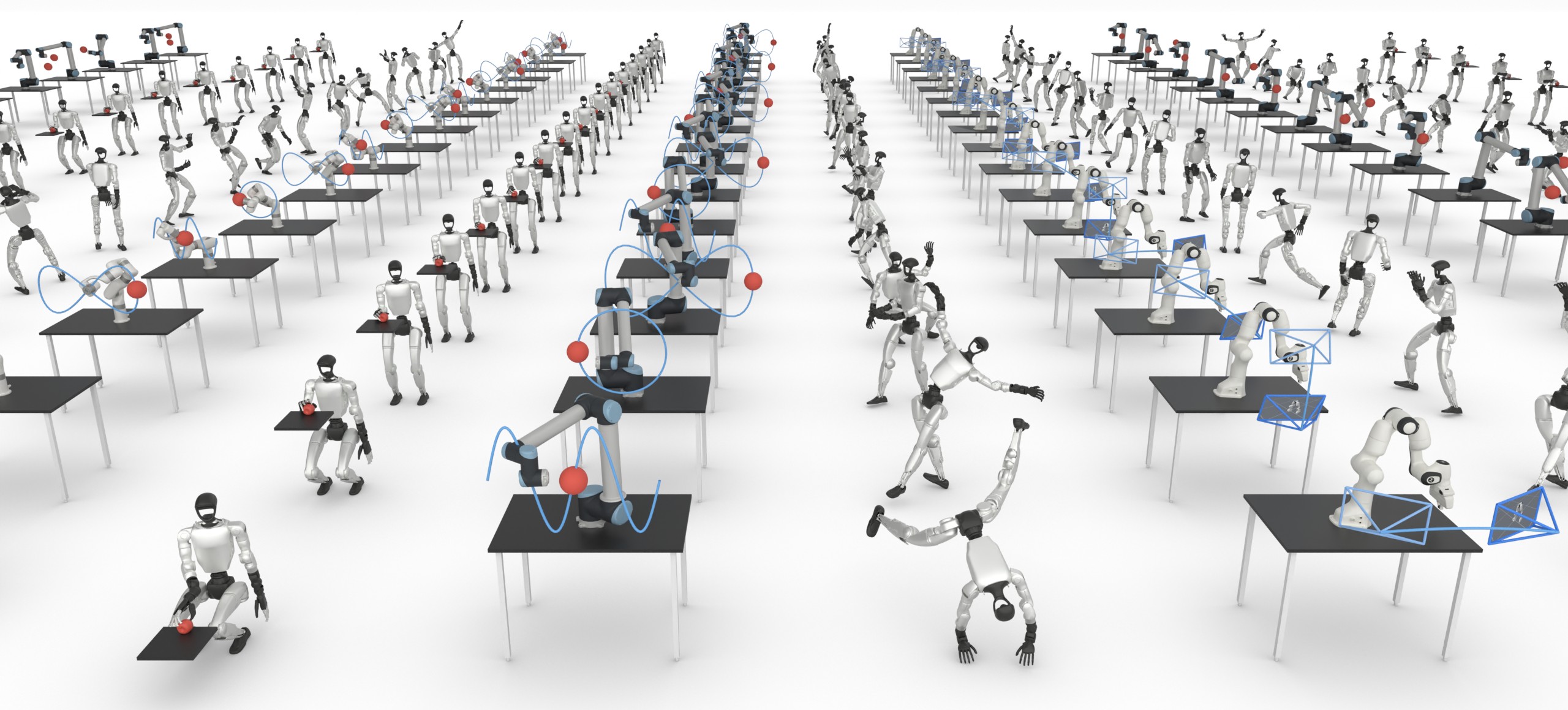}
    \vspace{-1em}
    \caption{\textbf{RoboKit} is a high-performance, general-purpose optimization toolkit for diverse robot tasks. We showcase representative examples; in the first row, from left to right: whole-body grasping, trajectory optimization, whole-body retargeting, and rendering-based calibration.}
    \label{fig:teaser}
\end{figure}

\begin{abstract}
    We present RoboKit, a high-performance, general-purpose toolkit for parallel differentiable optimization in robotics tasks, from inverse kinematics to simulation-based MPC. Its design follows three principles. (i) \emph{Performance and flexibility}: it implements GPU-parallel optimizers from first-order to sampling-based methods, together with composable cost terms for different tasks. With well-tuned engineering, it outperforms existing baselines with $2$--$60\times$ speedups. (ii) \emph{Geometry, rendering and physics}: it goes beyond kinematics to geometry, rendering and simulation, and accelerates them with hybrid and coarse-to-fine representations. For example, with hybrid geometry it optimizes 4096 grasps in 2.97\,s. (iii) \emph{Integration with learning}: with GPU-parallel simulation and differentiable cost terms, it supports policy distillation and reinforcement learning. The distilled policies run up to $240\times$ faster than optimization at comparable accuracy.
\end{abstract}

\keywords{Differentiable Optimization, Robot Kinematics, Policy Distillation}

\section{Introduction}

While today's robot learning embraces end-to-end policies trained at scale, classical optimization is quietly everywhere. It is invoked as a low-level skill during teleoperation and deployment~\cite{qin2023anyteleop,joao2025gmr,fu2026cap,kim2025pyroki,curobo_icra23}; it drives large-scale data generation~\cite{dexgraspnet,ye2025dex1b,yang2025deep,dalal2024neural,yang2025omniretarget}; it plans through physics in the loop~\cite{xue2025dial,pan2025spider}; and its objectives are distilled directly into policies~\cite{amos2017optnet,pineda2022theseus,yin2025geometric,zurbrugg2025graspqp}. In this paper, we ask how to better extend the value of optimization, accelerated by modern GPUs, to robot learning.

To this end, we present RoboKit, a performant, general-purpose toolkit supporting diverse tasks, including inverse kinematics (IK), trajectory optimization, motion retargeting, grasp synthesis, rendering-based calibration, and physics-based control. Three principles shape its design: (i) \textbf{Performance and flexibility}: it should be fast enough to keep pace with scaling, yet flexible enough to open new research directions; (ii) \textbf{Geometry, rendering and physics}: it should support geometry-aware tasks, differentiable rendering and simulation, which connect perception and action; and (iii) \textbf{Integration with learning}: its objectives and solvers should plug seamlessly into learning frameworks such as PyTorch.

\subhead{Performance and flexibility.} RoboKit implements a classical optimization framework in Warp~\cite{warp2022}, pairing multi-seed sampling with diverse optimizers: first-order, quasi-Newton, Gauss-Newton, and sampling-based methods. With well-tuned engineering, it solves over $10\times$ faster than PyRoKi~\cite{kim2025pyroki} on pure kinematic tasks. Unlike prior frameworks~\cite{curobo_icra23,kim2025pyroki}, these optimizers let us pick a suitable solver per task (Gauss-Newton for IK, L-BFGS for trajectory optimization), and a composable library of cost terms supports combined tasks such as geometry-aware retargeting.

\subhead{Geometry, rendering and physics.} Tasks beyond pure kinematics demand fine-grained geometry and differentiable rendering. RoboKit makes both first-class and accelerates them with voxel-SDF geometry, optional near-surface mesh refinement, and coarse-to-fine rendering. Using voxel SDFs, it synthesizes $4{,}096$ grasps in $2.97$\,s versus $178.45$\,s for Dex1B~\cite{ye2025dex1b} with higher success rate, and it calibrates $57\times$ faster than EasyHEC~\cite{chen2023easyhec} on its easy split (1.23 vs 70.4\,s), converging where EasyHEC fails. These capabilities are absent from prior frameworks. With physics in the loop, its MPPI~\cite{williams2017model} planner raises SPIDER's~\cite{pan2025spider} ADD-AUC from 51.3 to 65.8 on OakInk while running $5.6\times$ faster, and RL on the same physics reaches mjlab's~\cite{zakka2026mjlab} scores up to $2.1\times$ sooner.

\subhead{Integration with learning.} Every cost term is PyTorch-native and differentiable end to end, plugging directly into learning. We further study when a learned policy beats the solver and develop a general distillation framework combining offline and online learning. Policies also train with RL directly on RoboKit's parallel physics.

Our contributions are threefold:
\begin{itemize}[leftmargin=*,itemsep=2pt,topsep=0pt,parsep=0pt]
    \item A high-performance parallel optimization toolkit implemented in Warp, with the freedom to mix solvers and cost terms per task.
    \item Accelerated geometry, rendering and physics, extending optimization to contact-rich, rendering-based and physics-in-the-loop tasks beyond kinematics.
    \item Differentiable operators that plug into learning frameworks, with a general distillation framework for embedding optimization into policies.
\end{itemize}

\section{Related Work}

\noindent
\textbf{IK \& Trajectory Optimization.} Inverse kinematics seeks a joint configuration reaching a target end-effector pose. Early closed-form solvers~\cite{diankov2010automated,pieper1969kinematics} relied on restrictive assumptions, so recent work~\cite{beeson2015trac,mink,pink,kim2025pyroki} favors iterative optimization for its flexibility, support for secondary objectives, and collision handling across embodiments. Trajectory optimization~\cite{he2024learning,qin2023anyteleop,joao2025gmr,kim2025pyroki} extends this over time; since each timestep couples mainly to its neighbors, the resulting sparse problems benefit from sparse linear algebra.

\noindent
\textbf{Motion Retargeting.} Motion retargeting adapts existing motion data to new embodiments and is widely studied for both humanoid whole-body motion and dexterous manipulation. For humanoids, joint- and IK-based methods are common, but morphological differences cause artifacts such as foot sliding and penetration, motivating learning-based~\cite{aberman2020skeleton}, physics-augmented~\cite{won2019bodyshapevariation,ryu2021muscleretarget}, and contact-aware~\cite{villegas2021skinnedmotion,zhang2023multicharacter} formulations. Hand retargeting is more challenging due to dense contacts and highly articulated kinematics; approaches include direct joint mapping~\cite{rajeswaran2018dapg,kumar2015haptix}, keypoint-based IK~\cite{humbertson2015handson,qin2022from,antotsiou2018taskretargeting,dasari2023pgdm}, inverse dynamics~\cite{kry2006interactioncapture}, and functional pose equivalence~\cite{Handa2019DexPilotVT,sivakumar2022telekinesis}. Physics-based simulation and learning from demonstrations improve plausibility~\cite{wu2022ILAD} but are highly sensitive to retargeting quality, and standardized contact-aware methods remain lacking; SPIDER~\cite{pan2025spider} scales physics-based retargeting with sampling in simulation, while ManipTrans~\cite{li2025maniptrans} and DexMachina~\cite{mandi2025dexmachina} learn RL policies that transfer bimanual demonstrations.

\noindent
\textbf{Grasp Synthesis.} Dexterous grasp synthesis aims to generate stable grasp configurations for multi-fingered hands. Early approaches rely on analytic grasp metrics and optimization-based sampling with known object models~\cite{miller2004graspit, human-like-grasp}, but scale poorly to high-DoF hands. Learning-based methods have since become dominant: imitation learning trains on large-scale datasets from human demonstrations~\cite{taheri2020grab, liu2024realdex}, force-closure optimization~\cite{liu2021dfc, gendexgrasp}, or differentiable simulation~\cite{dexgraspnet}, enabling scalable supervision~\cite{unigrasp, gendexgrasp, contactgen}, while reinforcement learning learns closed-loop policies that are robust across objects~\cite{unidexgrasp, unidexgrasp++, graspgf, dexpoint}. BODex~\cite{chen2025bodex} scales optimization-based synthesis with GPU-parallel bilevel optimization. Still, scalable, human-like, and cross-embodiment grasp generation remains an open challenge~\cite{li2024multigraspllm}.

\noindent
\textbf{Optimization Libraries.} Existing toolkits each cover only part of the design space. CPU IK solvers such as TracIK~\cite{beeson2015trac}, pink~\cite{pink}, and mink~\cite{mink} stay IK-centered; TrajOpt~\cite{schulman2014motion} adds collision-aware motion generation; MoveIt~\cite{chitta2012moveit}, Drake~\cite{drake} and Crocoddyl~\cite{mastalli2020crocoddyl} offer general planning and optimal-control stacks on CPU, and VAMP~\cite{thomason2024vamp} vectorizes sampling-based planning; and GPU motion planners like cuRobo~\cite{curobo_icra23,sundaralingam2026curobov2} offer higher throughput. However, such motion-planning libraries typically target collision avoidance rather than contact-rich tasks like grasping. Sampling-based MPC frameworks such as MJPC~\cite{howell2022predictive} and DIAL-MPC~\cite{xue2025dial} optimize through physics, but target control alone. Retargeting tools~\cite{qin2023anyteleop,he2024learning} target hands or humanoids, while PyRoKi~\cite{kim2025pyroki} unifies these tasks into a single system across diverse robots, though it remains limited to kinematics-oriented tasks and primitive geometries. In contrast, RoboKit combines a GPU-parallel solver, accelerated geometry, rendering and physics, and PyTorch integration in a single framework, and is the only one to cover all of these tasks (Tab.~\ref{table:robokit_comparison}).

\begin{table}[t]
    \tablestyle
    \newcommand{\greencheck}{\textcolor[HTML]{1AB21A}{\scalebox{1.25}{\checkmark}}}
    \newcommand{\redx}{\textcolor[HTML]{E63333}{\scalebox{1.25}{\ding{55}}}}
    \caption{\textbf{Feature and task coverage} of representative optimization tools. \greencheck\greencheck denotes two-level coarse-to-fine support.}
    \label{table:robokit_comparison}
    \metricstable{l ccc ccccccc}{11}{%
            \toprule
             & \multicolumn{3}{c}{\textbf{Features}} & \multicolumn{7}{c}{\textbf{Tasks}} \\
            \cmidrule(lr){2-4} \cmidrule(lr){5-11}
            Method & \makecell{Hybrid\\Geometry} & Rendering & Physics & \makecell{Sim.\\MPC} & IK & \makecell{Traj.\\Opt.} & \makecell{Hand\\Retarget.} & \makecell{Humanoid\\Retarget.} & \makecell{Grasp\\Synth.} & \makecell{Render.\\Calib.} \\
            \midrule
            pink~\cite{pink}, mink~\cite{mink} & \redx & \redx & \redx & \redx & \greencheck & \redx & \redx & \redx & \redx & \redx \\
            TrajOpt~\cite{schulman2014motion} & \redx & \redx & \redx & \redx & \greencheck & \greencheck & \redx & \redx & \redx & \redx \\
            Dex-Retargeting~\cite{qin2023anyteleop} & \redx & \redx & \redx & \redx & \greencheck & \redx & \greencheck & \redx & \redx & \redx \\
            Geo-Retargeting~\cite{yin2025geometric} & \redx & \redx & \redx & \redx & \greencheck & \redx & \greencheck & \redx & \redx & \redx \\
            GMR~\cite{joao2025gmr} & \redx & \redx & \redx & \redx & \greencheck & \redx & \redx & \greencheck & \redx & \redx \\
            PyRoKi~\cite{kim2025pyroki} & \redx & \redx & \redx & \redx & \greencheck & \greencheck & \greencheck & \greencheck & \redx & \redx \\
            cuRobo~\cite{curobo_icra23} & \greencheck & \redx & \redx & \redx & \greencheck & \greencheck & \redx & \redx & \redx & \redx \\
            DIAL-MPC~\cite{xue2025dial} & \redx & \redx & \greencheck & \greencheck & \redx & \redx & \redx & \redx & \redx & \redx \\
            SPIDER~\cite{pan2025spider} & \redx & \redx & \greencheck & \greencheck & \redx & \redx & \greencheck & \greencheck & \redx & \redx \\
            \midrule
            \rowcolor{myblue}
            \textbf{RoboKit (Ours)} & \greencheck\greencheck & \greencheck\greencheck & \greencheck\greencheck & \greencheck & \greencheck & \greencheck & \greencheck & \greencheck & \greencheck & \greencheck \\
            \bottomrule

    }

\end{table}

\section{Method}

\begin{figure}[t]
    \centering
    \includegraphics[width=\textwidth]{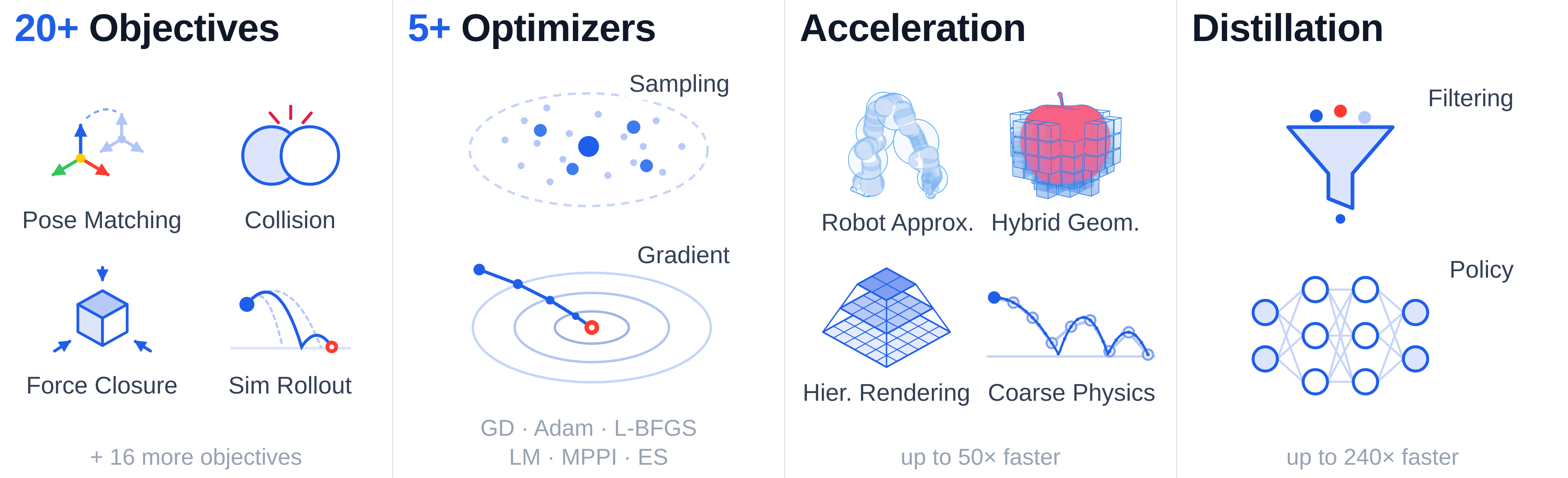}
    \caption{\textbf{Method Overview.} RoboKit has three components: a parallel \emph{solver} over differentiable objectives, an \emph{acceleration} structure for geometry, rendering and physics, and \emph{policy distillation} via offline and online learning.}
    \label{fig:method_ctrl}
    \vspace{-0.5em}
\end{figure}

RoboKit is built on three components: (i) a \emph{parallel solver} (Sec.~\ref{subsec:method1}) that pairs GPU-parallel multi-seed sampling with diverse optimizers (first-order, quasi-Newton, Gauss-Newton, and sampling-based) to produce high-quality solutions at scale; (ii) an \emph{accelerated structure} (Sec.~\ref{subsec:accel}) that makes geometry, differentiable rendering and physics first-class via hybrid and coarse-to-fine representations; and (iii) \emph{policy distillation} (Sec.~\ref{subsec:method2}) that distills optimization into fast policies. An overview of our framework is presented in Fig.~\ref{fig:method_ctrl}.

\subsection{Parallel Solver}
\label{subsec:method1}

\noindent
\textbf{Variables.} The core variable is the joint configuration $q \in \mathbb{R}^{d}$ ($d$ degrees of freedom), indexed over time as $q_t$ for trajectories. We support fixed, revolute, prismatic, and mimic joints, and $\mathrm{SE}(3)$/$\mathrm{SO}(3)$ Lie-group variables for poses. A differentiable forward kinematics maps $q$ and base pose ${}^{w}\mathbf{T}_{b}$ to joint poses ${}^{w}\mathbf{T}_{i} = \mathrm{FK}(q)_i$, with spatial or body Jacobians. We also allow auxiliary variables such as scale factors, common in retargeting to reduce the morphology gap.

\noindent
\textbf{Solver.} Our solver is task-agnostic and implemented in Warp, a GPU programming framework. It pairs (i) a high-level particle-based sampling solver with flexible sampling, gathering, distribution-fitting~\cite{hansen2016cma}, and resampling schedules, and (ii) a library of low-level optimizers, including first-order (gradient descent, Adam), quasi-Newton (limited-memory BFGS, L-BFGS), Gauss-Newton or Levenberg--Marquardt (LM), and sampling-based methods such as MPPI~\cite{williams2017model}, letting us pick the best solver per task (e.g., Gauss-Newton for IK, L-BFGS for trajectory optimization). Since Warp lacks native sparse Jacobians, we add coordinate-format (COO) sparse Jacobian support for all cost terms to scale to large variable dimensions. Curvature-based steps typically converge within 16--64 iterations; combined with large-scale parallel sampling, this yields a high-performance pipeline (Tab.~\ref{tab:metrics_abcde}).

\noindent
\textbf{Objectives.} We provide pre-built objective terms covering tasks from inverse kinematics to grasp synthesis, and support user-defined costs; analytic terms include analytical Jacobians, and the solver minimizes the sum of squared residuals. Terms can be combined, weighted, and reused, and also serve as supervision for online distillation. For example, the pose-matching cost penalizes the relative pose error $c_{\mathrm{pose}} = \log\!\left({}^{w}\mathbf{T}_{i}^{-1}\, {}^{w}\mathbf{T}_{\mathrm{tgt}}\right) \in \mathfrak{se}(3)$ via the logarithm map; see the supplementary for the full list. Objectives can also score physics rollouts in MuJoCo Warp~\cite{mujocowarp2025}, which sampling-based optimizers minimize without gradients.

\subsection{Accelerated Structure}
\label{subsec:accel}

Geometry, rendering and physics dominate the per-iteration cost of contact-rich, calibration and simulation tasks. We accelerate them with four structures that exploit spatial locality and a coarse-to-fine schedule.

\noindent
\textbf{Hybrid Voxel SDF and Mesh.} Querying a mesh SDF traverses a BVH and is expensive across thousands of sphere centers per iteration. A precomputed voxel SDF is cheap but its gradient is biased near the surface, where contact reasoning is most sensitive. We therefore pair each object $\mathcal{O}$ with a baked voxel SDF $\mathcal{V}$ (a grid padded by $\delta$ beyond its bounding box) and its source mesh $\mathcal{M}$, reading the cheap voxel field and refining to the exact mesh only inside a near-surface band:
$$
    \mathrm{SDF}({}^{w}c_i, \mathcal{O}) =
    \begin{cases}
        \mathrm{SDF}({}^{w}c_i, \mathcal{M}), & \mathrm{SDF}({}^{w}c_i, \mathcal{V}) \leq \tau, \\[2pt]
        \mathrm{SDF}({}^{w}c_i, \mathcal{V}), & \text{otherwise},
    \end{cases}
$$
with $\tau = b h$, where $h$ is the voxel size and $b$ is the configurable refinement-band width in voxels (default $b=4$). Far points cost an $O(1)$ voxel sample; near-contact points keep mesh-level accuracy and an exact normal $\hat{n}$.

\noindent
\textbf{Coarse-to-Fine Collision Queries.} The robot is approximated by per-link collision spheres (as in cuRobo~\cite{curobo_icra23}). We additionally wrap each link $\ell$ in one coarse bounding sphere (center ${}^{w}c_\ell$, radius $R_\ell$) that contains all of its collision spheres. Mirroring the hybrid query above, the effective distance $\widehat{\mathrm{SDF}}$ fed to the collision cost runs the exact per-sphere query only when the link is near the scene, and reuses the single broad-phase query otherwise:
$$
    \widehat{\mathrm{SDF}}({}^{w}c_i, \mathcal{O}) =
    \begin{cases}
        \mathrm{SDF}({}^{w}c_i, \mathcal{O}),    & \mathrm{SDF}({}^{w}c_\ell, \mathcal{O}) \leq R_\ell + \epsilon, \\[2pt]
        \mathrm{SDF}({}^{w}c_\ell, \mathcal{O}), & \text{otherwise}.
    \end{cases}
$$
A link with $\mathrm{SDF}({}^{w}c_\ell, \mathcal{O}) > R_\ell + \epsilon$ keeps all of its spheres beyond the safety margin $\epsilon$, so the cost is inactive and one broad query prunes the whole link, which is especially effective for hands and articulated arms.

\noindent
\textbf{Coarse-to-Fine Rendering.} Rendering-based calibration matches rendered robot masks to observations, a non-convex silhouette objective. We use a resolution pyramid: early stages render at low resolution with many seeds and a strongly blurred, smoother cost with fewer local minima so sampling escapes poor basins cheaply; later stages raise resolution, prune to the best seeds via beam selection, and remove the blur for sub-pixel refinement. A single-level full-resolution solver fails under large initialization error, whereas the pyramid converges to sub-degree accuracy.

\noindent
\textbf{Coarse-to-Fine Physics.} Sampling-based MPC and RL spend most of their time stepping the simulator. Early iterations run a coarse scene model, with larger timesteps, fewer solver iterations and fewer collision pairs; later ones switch to full physics.

\subsection{Policy Learning}
\label{subsec:method2}

Policy distillation transfers the solver's knowledge to a policy through two complementary paradigms: \emph{offline} and \emph{online} distillation.

\noindent
\textbf{Offline Distillation.}
We run the solver over a distribution of task instances to generate solved configurations. Since poorly converged or infeasible solutions degrade learning, we optionally filter to keep only high-quality demonstrations, e.g., by thresholding the optimization cost or verifying physical plausibility in simulation; this is critical for hard tasks like grasping. Given the filtered dataset $\mathcal{D} = \{(o_i, q_i^*)\}$, we train a policy $\pi_\theta(o)$ via supervised learning:
$$
    \mathcal{L}_{\mathrm{offline}}(\theta) = \mathbb{E}_{(o, q^*) \sim \mathcal{D}} \left[ \left\| \pi_\theta(o) - q^* \right\|^2 \right].
$$

\noindent
\textbf{Online Distillation.}
Online distillation trains the policy directly from optimization objectives without a fixed dataset, backpropagating gradients to the policy weights:
$$
    \mathcal{L}_{\mathrm{online}}(\theta) = \mathbb{E}_{o \sim p(o)} \left[ \frac{1}{2}\sum_k \bigl\|W_k \, c_k\bigl(\pi_\theta(o); o\bigr)\bigr\|_2^2 \right],
$$
where $c_k(q; o)$ is the $k$-th unweighted residual vector and $W_k$ is its diagonal residual-weight matrix, including the task gain. Gradients flow through these differentiable residuals to the policy; the solver is never unrolled or differentiated. Operating on-policy, it provides exact gradients and draws fresh samples each iteration, typically outperforming offline distillation and generalizing better.

\noindent
\textbf{Integrating Offline and Online.}
Online distillation performs better but depends on task complexity. For simple tasks like position-only retargeting it suffices alone; for contact-rich tasks like grasp synthesis, pure online training converges to poor local minima, so we warm up with offline supervised learning on filtered demonstrations before online fine-tuning. This hybrid recipe mirrors imitation and reinforcement learning practice, but benefits from the stable gradients of differentiable kinematic objectives.

\section{Experiments}

We address three questions in experiments: (1) How well does our optimization framework perform? (2) How effective are policies learned with RoboKit? (3) Which design choices matter? We answer these questions below.

\subsection{Experimental Settings}

\noindent
We evaluate RoboKit on six kinematic tasks and one physics-based task, reusing established benchmarks: the cuRobo~\cite{curobo_icra23} benchmark for inverse kinematics and trajectory optimization (7-DoF Franka Panda), Dex-Retargeting~\cite{qin2023anyteleop} data for hand motion retargeting (Shadow Hand), AMASS~\cite{mahmood2019amass} for human motion retargeting (Unitree G1), Dex1B~\cite{ye2025dex1b} for grasp synthesis (MANO on DexGraspNet~\cite{dexgraspnet}), the EasyHEC~\cite{chen2023easyhec} benchmark for rendering-based calibration (Panda), and SPIDER's~\cite{pan2025spider} MuJoCo comparison for physics-based dexterous retargeting (five hands on OakInk and GigaHands), and mjlab's~\cite{zakka2026mjlab} RL tasks for policy learning (Go1, G1 and a Yam arm), a GPU framework in the line of Isaac Gym~\cite{makoviychuk2021isaac} and MuJoCo Playground~\cite{zakka2025mujoco}. We report solve time alongside the task-specific success and accuracy metrics defined in each table. All benchmark comparisons (Tab.~\ref{tab:metrics_abcde}) run on a single NVIDIA RTX 4090 GPU; timings are averaged over 5 runs (3 for human retargeting, one pass over all frames for hand retargeting) after a warmup.

\begin{table*}[t]
    \tablestyle
    \caption{\textbf{Quantitative comparison} of RoboKit against baselines across seven tasks. See the appendix for metric definitions.}
    \label{tab:metrics_abcde}
    \begin{subtable}[t]{\textwidth}
        \centering
        \caption{Batched IK on the 7-DoF Franka Panda.}
        \label{tab:metric_a_ik}
        \metricstable{l cccc cccc}{9}{%
                \toprule
                      & \multicolumn{4}{c}{PyRoKi~\cite{kim2025pyroki}} & \multicolumn{4}{c}{Ours (opt)} \\
                \cmidrule(lr){2-5} \cmidrule(lr){6-9}
                Batch & Time (ms) & Succ. (\%) & Pos. Err (mm) & Ori. Err (rad) & Time (ms) & Succ. (\%) & Pos. Err (mm) & Ori. Err (rad) \\
                \midrule
                1 & 9.08 & 100.00 & $1.74\times10^{-4}$ & $1.88\times10^{-7}$ & \cellcolor{myblue}\textbf{0.65} & \cellcolor{myblue}100.00 & \cellcolor{myblue}$1.23\times10^{-4}$ & \cellcolor{myblue}$1.68\times10^{-7}$ \\
                10 & 14.52 & 100.00 & $3.80\times10^{-2}$ & $4.86\times10^{-5}$ & \cellcolor{myblue}\textbf{0.61} & \cellcolor{myblue}100.00 & \cellcolor{myblue}$1.15\times10^{-2}$ & \cellcolor{myblue}$1.90\times10^{-5}$ \\
                100 & 16.04 & 100.00 & $1.91\times10^{-2}$ & $2.13\times10^{-5}$ & \cellcolor{myblue}\textbf{0.81} & \cellcolor{myblue}100.00 & \cellcolor{myblue}$1.08\times10^{-3}$ & \cellcolor{myblue}$1.20\times10^{-6}$ \\
                1000 & 48.19 & 99.88 & $1.47\times10^{-3}$ & $1.01\times10^{-6}$ & \cellcolor{myblue}\textbf{3.41} & \cellcolor{myblue}99.94 & \cellcolor{myblue}$4.04\times10^{-4}$ & \cellcolor{myblue}$4.15\times10^{-7}$ \\
                2000 & 86.20 & 99.89 & $1.57\times10^{-3}$ & $1.05\times10^{-6}$ & \cellcolor{myblue}\textbf{7.12} & \cellcolor{myblue}99.91 & \cellcolor{myblue}$3.28\times10^{-4}$ & \cellcolor{myblue}$4.03\times10^{-7}$ \\
                \bottomrule
        }
    \end{subtable}
    \par\vspace{4pt}\nointerlineskip

    \begin{subtable}[t]{\textwidth}
        \centering
        \caption{Motion Planning for the Franka arm.}
        \label{tab:metric_b_trajopt}
        \metricstable{lcccccccc}{9}{%
            \toprule
            Method & \makecell{Success\\(\%) $\uparrow$} & \makecell{Plan Mean\\(ms) $\downarrow$} & \makecell{Plan P98\\(ms) $\downarrow$} & \makecell{Motion Time\\(s) $\downarrow$} & \makecell{Path Length\\(rad) $\downarrow$} & \makecell{Max Jerk\\(rad/s$^3$) $\downarrow$} & \makecell{Mean Vel\\(rad/s)} & \makecell{Max Acc\\(rad/s$^2$) $\downarrow$} \\
            \midrule
            cuRobo~\cite{curobo_icra23} & 99.57 & 56.6 & 214.5 & \textbf{1.181} & 3.145 & 146.2 & 0.698 & \textbf{11.567} \\
            \rowcolor{myblue}
            Ours (opt) & \textbf{99.65} & \textbf{16.3} & \textbf{40.5} & 1.189 & \textbf{3.131} & \textbf{118.1} & 0.766 & 11.941 \\
            \bottomrule
        }
    \end{subtable}
    \par\vspace{4pt}\nointerlineskip

    \begin{subtable}[t]{0.49\textwidth}
        \centering
        \caption{Online hand retargeting for Shadow Hand.}
        \label{tab:metric_c_handret}
            \metricstable{lccccc}{6}{%
                \toprule
                Method & \makecell{Time\\(ms) $\downarrow$} & \makecell{Pos.\\(mm) $\downarrow$} & \makecell{Pinch\\(\%) $\uparrow$} & \makecell{Open\\(\%) $\uparrow$} & \makecell{Acc.\\(rad/s$^2$) $\downarrow$} \\
                \midrule
                Dex-Ret. (pos)~\cite{qin2023anyteleop}      & 3.66                   & 28.13                  & 37.5                  & 65.2                 & 7.44                           \\
                Dex-Ret. (DexPilot)~\cite{qin2023anyteleop} & 15.71                  & 15.69                  & 96.9                  & 85.2                 & \textbf{4.23}                  \\
                PyRoKi~\cite{kim2025pyroki}                 & 33.19                  & 10.76                  & 35.7                  & \textbf{94.3}        & 31.02                          \\
                \rowcolor{myblue}
                Ours (opt)                                  & \textbf{1.25}          & \textbf{6.05}          & \textbf{99.7}         & 92.3                 & 5.82                           \\
                \bottomrule
            }
    \end{subtable}
    \hfill
    \begin{subtable}[t]{0.49\textwidth}
        \centering
        \caption{Online humanoid retargeting for G1.}
        \label{tab:metric_d_humanret}
            \metricstable{lcccc}{5}{%
                \toprule
                Method & \makecell{Time\\(ms) $\downarrow$} & \makecell{MPJPE\\(cm) $\downarrow$} & \makecell{WA-MPJPE\\(cm) $\downarrow$} & \makecell{Acc.\\(rad/s$^2$) $\downarrow$} \\
                \midrule
                GMR~\cite{joao2025gmr} & 7.91                   & 9.26                     & 9.21                        & 20.06                          \\
                \rowcolor{myblue}
                Ours (opt)             & \textbf{2.00}          & \textbf{8.08}            & \textbf{5.57}               & \textbf{16.98}                 \\
                \bottomrule
            }
    \end{subtable}
    \par\vspace{4pt}\nointerlineskip

    \begin{subtable}[t]{0.49\textwidth}
        \centering
        \caption{Grasp synthesis.}
        \label{tab:metric_e_grasp}
            \metricstable{lcccc}{5}{%
                \toprule
                Method & \makecell{Time\\(s) $\downarrow$} & \makecell{SR\\(\%) $\uparrow$} & \makecell{Q1\\$\uparrow$} & \makecell{Pen.\\(mm) $\downarrow$} \\
                \midrule
                Dex1B~\cite{ye2025dex1b} & 178.45                & 23.63              & \textbf{0.109} & 1.06                   \\
                \rowcolor{myblue}
                Ours (opt)               & \textbf{2.97}         & \textbf{46.83}     & 0.077          & \textbf{0.70}          \\
                \bottomrule
            }
    \end{subtable}
    \hfill
    \begin{subtable}[t]{0.49\textwidth}
        \centering
        \caption{Hand-eye calibration on Panda (realistic).}
        \label{tab:metric_f_calib}
            \metricstable{lccc}{4}{%
                \toprule
                Method & \makecell{Time\\(s) $\downarrow$} & \makecell{Trans.\\(cm) $\downarrow$} & \makecell{Rot.\\($^\circ$) $\downarrow$} \\
                \midrule
                EasyHEC~\cite{chen2023easyhec} & 458.5                 & 96.43                    & 12.66                        \\
                \rowcolor{myblue}
                Ours (opt)                     & \textbf{2.0}          & \textbf{1.10}            & \textbf{0.54}                \\
                \bottomrule
            }
    \end{subtable}
    \par\vspace{4pt}\nointerlineskip

    \begin{subtable}[t]{\textwidth}
        \centering
        \caption{Physics-based dexterous retargeting on five hands (SPIDER's MuJoCo comparison); time per frame.}
        \label{tab:metric_g_physret}
        \metricstable{l ccccc ccccc}{11}{%
            \toprule
             & \multicolumn{5}{c}{OakInk (143 trials)} & \multicolumn{5}{c}{GigaHands (75 trials)} \\
            \cmidrule(lr){2-6} \cmidrule(lr){7-11}
            Method & \makecell{Time\\(ms) $\downarrow$} & \makecell{ADD-AUC\\(\%) $\uparrow$} & \makecell{Succ.\\(\%) $\uparrow$} & \makecell{Pos.\\(cm) $\downarrow$} & \makecell{Rot.\\($^\circ$) $\downarrow$} & \makecell{Time\\(ms) $\downarrow$} & \makecell{ADD-AUC\\(\%) $\uparrow$} & \makecell{Succ.\\(\%) $\uparrow$} & \makecell{Pos.\\(cm) $\downarrow$} & \makecell{Rot.\\($^\circ$) $\downarrow$} \\
            \midrule
            SPIDER~\cite{pan2025spider} & 610 & 51.3 & 59 & 4.38 & 34.8 & 125 & 50.3 & 77 & \textbf{4.05} & 16.0 \\
            \rowcolor{myblue}
            Ours (opt) & \textbf{109} & \textbf{65.8} & \textbf{86} & \textbf{3.17} & \textbf{15.2} & \textbf{57} & \textbf{63.7} & \textbf{84} & 4.11 & \textbf{9.5} \\
            \bottomrule
        }
    \end{subtable}

\end{table*}

\subsection{Parallel Solver Performance}
\begin{figure*}
    \centering
    \captionsetup{type=figure}
    \includegraphics[width=\linewidth]{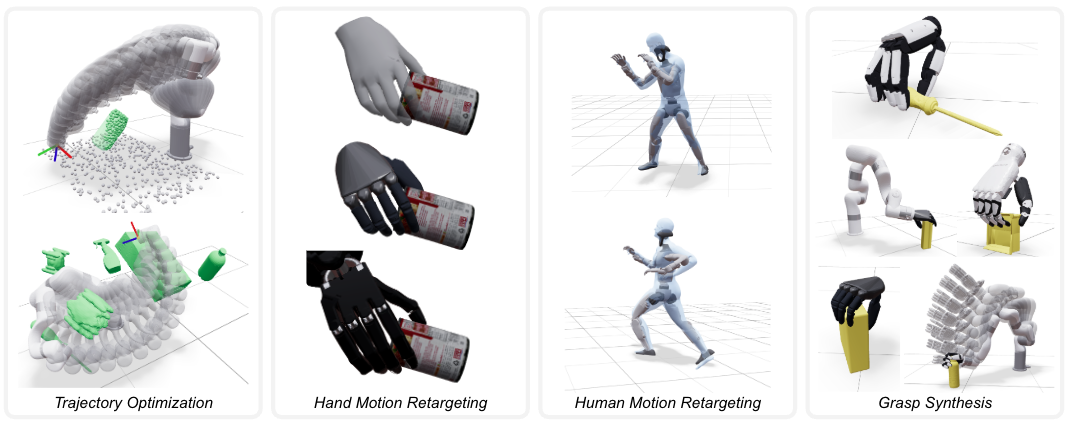}
    \vspace{-1em}
    \caption{\textbf{Qualitative results} of the learned policies on trajectory optimization, hand and full-body retargeting, and dexterous grasp synthesis.}
    \label{fig:tasks}
\end{figure*}

We compare RoboKit against optimization baselines across all seven tasks (Tab.~\ref{tab:metrics_abcde}); it is consistently the fastest, with task-specific tradeoffs in accuracy and solution quality.

\noindent
\emph{(a) Inverse Kinematics.} Against PyRoKi~\cite{kim2025pyroki} both methods use Levenberg--Marquardt, $64$ initial seeds refined to $4$, and a $16$-iteration budget. The resulting speedup (e.g., $7.1$ vs $86.2$\,ms at batch $2000$) reflects implementation and solver-setting differences: our Warp~\cite{warp2022} backend removes the per-iteration framework overhead of JAX, especially at large batches, while reaching the same near-$100\%$ success and comparable accuracy.

\noindent
\emph{(b) Trajectory Optimization.} We adopt the strong cuRobo~\cite{curobo_icra23} benchmark and, like it, optimize with L-BFGS. Both methods are judged on goal reach, world and self-collision, dynamic limits, and joint limits with the benchmark's $0.2$\,rad tolerance. RoboKit is faster ($0.016$ vs $0.057$\,s plan time, and a far lower $98$th-percentile time), maintains a comparable success rate, and yields a shorter joint path and lower jerk, at the cost of slightly longer motion time. We omit PyRoKi here: its LM solver is prohibitively slow in this setting, and it performs trajectory optimization without time reparameterization, so its motion-level metrics are not directly comparable. Note that the obstacles here are primitive boxes, so our hybrid geometry acceleration (Sec.~\ref{subsec:accel}) is not exercised; the gains come purely from the solver and engineering.

\noindent
\emph{(c) Hand Retargeting.} Dex-Retargeting~\cite{qin2023anyteleop} optimizes a single objective at a time (position \emph{or} vector), whereas RoboKit and PyRoKi jointly satisfy both. Even against these single-objective specialists, RoboKit attains lower position error, and is $26.6\times$ faster than PyRoKi at a higher pinch success rate.

\noindent
\emph{(d) Human Retargeting.} Against GMR~\cite{joao2025gmr}, we retarget human motion from the AMASS dataset~\cite{mahmood2019amass} to the G1 humanoid; RoboKit cuts the per-frame time by about $4\times$ ($2.00$ vs $7.91$\,ms) while lowering both MPJPE and WA-MPJPE.

\noindent
\emph{(e) Grasp Synthesis.} RoboKit optimizes $4{,}096$ grasps using voxel-SDF geometry in $2.97$\,s, versus $178.45$\,s for Dex1B~\cite{ye2025dex1b} ($60\times$ faster). Optimization uses a precomputed voxel field; all grasps are scored using the object mesh. It has higher success ($46.83\%$ vs $23.63\%$) and lower penetration ($0.70$ vs $1.06$\,mm), while Dex1B achieves higher Q1 ($0.109$ vs $0.077$). Mesh results and the geometry ablation appear in Tab.~\ref{tab:grasp_run_statistics} and Tab.~\ref{tab:ce_geom}.

\noindent
\emph{(f) Rendering-Based Calibration.} We evaluate the realistic setting matching our introduction ($35$\,cm/$45^\circ$ initial error with sensor and calibration noise). The coarse-to-fine pyramid (Sec.~\ref{subsec:accel}) reaches $1.10$\,cm in $2.0$\,s, while EasyHEC~\cite{chen2023easyhec} does not converge ($96.43$\,cm). On the easy split ($10$\,cm/$15^\circ$ initial error, no added noise), where both reach under $0.3$\,cm, we are $57\times$ faster ($1.23$\,s vs $70.4$\,s).

\noindent
\emph{(g) Physics-Based Retargeting.} Following SPIDER~\cite{pan2025spider}, we retarget human hand--object motion to five dexterous hands by sampling-based MPC in MuJoCo Warp~\cite{todorov2012mujoco,mujocowarp2025}, on $7$ OakInk~\cite{zhan2024oakink2} tasks and $3$ GigaHands~\cite{fu2025gigahands} clips with five seeds each; only the optimizer differs from SPIDER. RoboKit raises ADD-AUC from $51.3$ to $65.8$ on OakInk and from $50.3$ to $63.7$ on GigaHands, and is $5.6\times$ and $2.2\times$ faster.

\subsection{Policy Learning Performance}

We evaluate policies distilled across four tasks, excluding inverse kinematics, which optimization already solves trivially; results are in Tab.~\ref{tab:neural} (distilled rows in blue) and Fig.~\ref{fig:tasks}. Overall, distilled policies run $9$--$240\times$ faster than optimization at competitive accuracy, enabling real-time deployment. All methods use offline training.

\begin{table*}[t]
    \tablestyle
    \caption{\textbf{Distilled policy performance} across four tasks.}
    \label{tab:neural}
    \begin{subtable}[t]{0.49\textwidth}
        \centering
        \caption{Trajectory optimization.}
        \label{tab:neural_traj}
        \metricstable{lccccc}{6}{%
                \toprule
                Method & \makecell{Time\\(ms) $\downarrow$} & \makecell{SR\\(\%) $\uparrow$} & \makecell{Pos.\\(cm)} & \makecell{Rot.\\($^\circ$)} & Smooth. \\
                \midrule
                Ours (opt)    & 150.61                 & \textbf{96.72}     & \textbf{1.04} & \textbf{1.49}   & \textbf{0.256} \\
                \rowcolor{myblue}
                Ours (distilled) & \textbf{0.63}          & 95.90              & 1.08          & 2.32            & 0.588          \\
                \bottomrule

    }
    \end{subtable}
    \hfill
    \begin{subtable}[t]{0.49\textwidth}
        \centering
        \caption{Hand retargeting.}
        \label{tab:neural_hand}
        \metricstable{lcccc}{5}{%
                \toprule
                Method & \makecell{Time\\(ms) $\downarrow$} & \makecell{SR\\(\%) $\uparrow$} & \makecell{Pos.\\(mm) $\downarrow$} & \makecell{Vec.\\(mm) $\downarrow$} \\
                \midrule
                Ours (opt)    & 5.22                   & 99.72              & 7.79                   & 8.81                   \\
                \rowcolor{myblue}
                Ours (distilled) & \textbf{0.44}          & \textbf{100.0}     & \textbf{4.80}          & \textbf{6.5}           \\
                \bottomrule

    }
    \end{subtable}
    \par\vspace{4pt}\nointerlineskip

    \begin{subtable}[t]{0.49\textwidth}
        \centering
        \caption{Humanoid retargeting.}
        \label{tab:neural_human}
        \metricstable{lccc}{4}{%
                \toprule
                Method & \makecell{Time\\(ms) $\downarrow$} & \makecell{Pos.\\(cm) $\downarrow$} & \makecell{Ori.\\(rad) $\downarrow$} \\
                \midrule
                Ours (opt)    & 10.86                  & \textbf{7.15}          & \textbf{0.1613}         \\
                \rowcolor{myblue}
                Ours (distilled) & \textbf{0.72}          & 9.92                   & 0.1936                  \\
                \bottomrule

    }
    \end{subtable}
    \hfill
    \begin{subtable}[t]{0.49\textwidth}
        \centering
        \caption{Grasp synthesis.}
        \label{tab:neural_grasp}
        \metricstable{lcccc}{5}{%
                \toprule
                Method & \makecell{Time\\(ms) $\downarrow$} & \makecell{SR\\(\%) $\uparrow$} & \makecell{Q1\\$\uparrow$} & \makecell{Pen.\\(cm) $\downarrow$} \\
                \midrule
                UGG~\cite{lu2024ugg}     & --                     & 43.6               & 0.026          & 0.43              \\
                Dex1B~\cite{ye2025dex1b} & 1.71                   & 63.7               & 0.075          & 0.29              \\
                Ours (opt)               & 14.87                  & 31.25              & \textbf{0.110} & 0.99              \\
                \rowcolor{myblue}
                Ours (distilled)         & \textbf{1.71}          & \textbf{71.3}      & 0.089          & \textbf{0.18}     \\
                \bottomrule

    }
    \end{subtable}

\end{table*}

\begin{table}[t]
    \tablestyle
    \caption{\textbf{RL training} against mjlab~\cite{zakka2026mjlab} on the same MuJoCo Warp physics: minutes to reach $95\%$ / $100\%$ of mjlab's final score, median over seeds.}
    \label{tab:rl_mjlab}
    \metricstable{l cccccc}{7}{%
        \toprule
        Method & \makecell{Go1\\Flat} & \makecell{G1\\Flat} & \makecell{Go1\\Rough} & \makecell{G1\\Rough} & \makecell{G1\\Tracking} & \makecell{Yam\\Lift} \\
        \midrule
        mjlab~\cite{zakka2026mjlab} & 7.6 / 37.6 & 18.5 / 47.2 & 20.0 / 45.6 & 77.6 / 102.1 & 49.0 / 51.1 & 7.4 / 16.8 \\
        \rowcolor{myblue}
        Ours & \textbf{3.2} / \textbf{17.7} & \textbf{9.2} / \textbf{26.7} & \textbf{12.7} / \textbf{34.4} & \textbf{46.4} / \textbf{74.6} & \textbf{21.9} / \textbf{29.3} & \textbf{3.8} / \textbf{8.7} \\
        \bottomrule
    }

\end{table}

Policies vs.\ our optimizer, and vs.\ learned grasping baselines (UGG, Dex1B). Trajectory and human-retargeting times are per frame. Optimizer rows follow each policy's evaluation protocol on an RTX A5000, which differs from Tab.~\ref{tab:metrics_abcde}.

Per-task results follow. \emph{(a) Trajectory optimization:} the policy amortizes the solver into a $239\times$ faster forward pass ($0.63$ vs.\ $150.61$\,ms/frame) at matched success, with only a slight smoothness drop. \emph{(b) Hand retargeting:} sub-millisecond inference that also \emph{improves} accuracy over the optimizer (position error $7.79\!\rightarrow\!4.80$\,mm), as in-distribution training escapes the local minima the optimizer settles into. \emph{(c) Human retargeting:} a $15\times$ speedup ($0.72$ vs.\ $10.86$\,ms/frame) with only a modest accuracy drop. \emph{(d) Grasp synthesis:} our distilled policy achieves higher success and lower penetration than UGG~\cite{lu2024ugg}, Dex1B~\cite{ye2025dex1b}, and the optimizer, with competitive grasp quality. Beyond distillation, RoboKit trains RL policies on mjlab's tasks with the same physics (Tab.~\ref{tab:rl_mjlab}), reaching mjlab's final score $1.3$--$2.1\times$ sooner.

\begin{table}[t]
    \tablestyle
    \caption{\textbf{Ablation study} of distillation, demonstration filtering, and geometry.}
    \label{tab:metrics_ce}
    \begin{subtable}[t]{0.49\textwidth}
        \centering
        \caption{Offline vs.\ online (hand retargeting).}
        \label{tab:ce_hand}
        \metricstable{lccc}{4}{%
                \toprule
                Method & \makecell{SR\\(\%) $\uparrow$} & \makecell{Pos. Err\\(mm) $\downarrow$} & \makecell{Vec. Err\\(mm) $\downarrow$} \\
                \midrule
                Offline & 100.0              & 4.71                       & 7.86                       \\
                Online (Supervised)     & 100.0              & 4.34                       & 6.87                       \\
                Online (Direct Grad.) & 100.0              & \textbf{2.71}              & \textbf{6.59}              \\
                \bottomrule

    }
    \end{subtable}
    \hfill
    \begin{subtable}[t]{0.49\textwidth}
        \centering
        \caption{Offline vs.\ hybrid (grasp).}
        \label{tab:ce_grasp}
        \metricstable{lccc}{4}{%
                \toprule
                Method & \makecell{SR\\(\%) $\uparrow$} & \makecell{Q1\\$\uparrow$} & \makecell{Pen. Err\\(mm) $\downarrow$} \\
                \midrule
                Offline          & 57.91              & 0.118         & 3.51                       \\
                Offline + Online & \textbf{65.48}     & \textbf{0.126} & \textbf{2.13}                      \\
                \bottomrule

    }
    \end{subtable}
    \par\vspace{4pt}\nointerlineskip

    \begin{subtable}[t]{0.49\textwidth}
        \centering
        \caption{Demonstration filtering (grasp).}
        \label{tab:ce_filter}
        \metricstable{lccc}{4}{%
                \toprule
                Method & \makecell{SR\\(\%) $\uparrow$} & \makecell{Q1\\$\uparrow$} & \makecell{Pen. Err\\(mm) $\downarrow$} \\
                \midrule
                Offline (w/o. filtering) & 23.83              & 0.079         & 6.82                       \\
                Offline                  & \textbf{57.91}     & \textbf{0.118} & \textbf{3.51}                      \\
                \bottomrule

    }
    \end{subtable}
    \hfill
    \begin{subtable}[t]{0.49\textwidth}
        \centering
        \caption{Geometry (grasp): voxel-only, scored on the mesh.}
        \label{tab:ce_geom}
        \metricstable{lccc}{4}{%
                \toprule
                Geometry & \makecell{Time\\(s) $\downarrow$} & \makecell{SR\\(\%) $\uparrow$} & \makecell{Pen.\\(mm) $\downarrow$} \\
                \midrule
                Mesh SDF (BVH)          & 16.61                 & \textbf{48.49}     & \textbf{0.55}          \\
                Voxel SDF (Ours)        & \textbf{2.97}         & 46.83              & 0.70          \\
                \bottomrule

    }
    \end{subtable}

\end{table}

\subsection{Ablation Study}

\noindent
\textbf{Offline vs.\ Online Distillation.}
Online (Supervised) learns fresh RoboKit-optimized targets, whereas Online (Direct Grad.) directly minimizes the task objective. On the simpler hand-retargeting task, pure online distillation wins, as on-policy sampling yields more accurate gradients (Tab.~\ref{tab:metrics_ce}a); here we sample random hand configurations rather than DexYCB motions so the gap is visible. On the harder grasp task, pure online training diverges in the non-convex landscape, so we warm-start offline and fine-tune online; this hybrid recipe yields the best success, quality, and penetration (Tab.~\ref{tab:metrics_ce}b).

\noindent
\textbf{Effect of Demonstration Filtering.}
Training offline on unfiltered solver outputs underperforms even the optimizer; simulation-based filtering to keep only successful grasps roughly doubles success and nearly halves penetration (Tab.~\ref{tab:metrics_ce}c), confirming that data quality is critical for contact-rich tasks.

\begin{wraptable}{r}{0.44\linewidth}
    \vspace{-4pt}
    \tablestyle
    \caption{\textbf{Solver component ablation} on the 7-DoF Franka Panda.}
    \label{tab:solver_ablation}
    \metricstable{l cc}{3}{%
        \toprule
        Variant & \makecell{Time\\(ms) $\downarrow$} & \makecell{SR\\(\%) $\uparrow$} \\
        \midrule
        Full           & \textbf{1.82}          & \textbf{100.0}     \\
        w/o Warp       & 139.59                 & 100.0              \\
        w/o Anal.\ Jac & 17.93                  & 100.0              \\
        w/o Particle   & 1.64                   & 72.6               \\
        \bottomrule

    }

\vspace{-10pt}
\end{wraptable}
\noindent
\textbf{Geometry Representation.}
Holding the solver fixed, replacing exact mesh-BVH queries with a precomputed voxel SDF cuts grasp-optimization time about $5.6\times$ at comparable success and negligible penetration change (Tab.~\ref{tab:metrics_ce}d).

\noindent
\textbf{Solver Components.} Table~\ref{tab:solver_ablation} isolates the solver's components on Franka IK. PyTorch kernels slow solves $77\times$ ($1.82\to139.59$\,ms) and numerical Jacobians $10\times$, both at $100\%$ success; removing multi-seed sampling barely saves time but drops success to $72.6\%$.

\vspace{-5pt}
\section{Conclusion}
\vspace{-2pt}
We presented RoboKit, a high-performance, general-purpose toolkit for parallel differentiable optimization in robotics. Built in Warp with versatile optimizers, it runs $2$--$60\times$ faster than task-specific baselines; it makes geometry, differentiable rendering and physics first-class; and its differentiable costs and parallel simulation plug into learning, distilling optimization into policies that run $9$--$240\times$ faster at comparable accuracy. We hope RoboKit offers a practical foundation for real-time manipulation, teleoperation, and learning-from-demonstration.

\vspace{-5pt}
\section{Limitations}
\vspace{-2pt}
RoboKit targets GPU-parallel workloads: for single-instance kinematics on CPU, kernel-launch overhead dominates and a C++ solver would be faster. Its trajectory optimization has no graph-search stage, which current benchmarks do not stress, and it lacks capabilities such as inverse dynamics.

\section*{Acknowledgment}
This work was supported, in part, by NSF CAREER Award IIS-2240014 and NSF CCF-2112665 (TILOS).

\bibliography{references}

\begin{thebibliography}{81}
\providecommand{\natexlab}[1]{#1}
\providecommand{\url}[1]{\texttt{#1}}
\expandafter\ifx\csname urlstyle\endcsname\relax
  \providecommand{\doi}[1]{doi: #1}\else
  \providecommand{\doi}{doi: \begingroup \urlstyle{rm}\Url}\fi

\bibitem[Qin et~al.(2023)Qin, Yang, Huang, Van~Wyk, Su, Wang, Chao, and
  Fox]{qin2023anyteleop}
Y.~Qin, W.~Yang, B.~Huang, K.~Van~Wyk, H.~Su, X.~Wang, Y.-W. Chao, and D.~Fox.
\newblock Anyteleop: A general vision-based dexterous robot arm-hand
  teleoperation system.
\newblock In \emph{RSS}, 2023.

\bibitem[Araujo et~al.(2026)Araujo, Ze, Xu, Wu, and Liu]{joao2025gmr}
J.~P. Araujo, Y.~Ze, P.~Xu, J.~Wu, and C.~K. Liu.
\newblock Retargeting matters: General motion retargeting for humanoid motion
  tracking.
\newblock In \emph{IEEE International Conference on Robotics and Automation
  (ICRA)}, 2026.

\bibitem[Fu et~al.(2026)Fu, Yu, El-Refai, Kou, Xue, Huang, Xiao, Wang, Li, Shi,
  et~al.]{fu2026cap}
M.~Fu, J.~Yu, K.~El-Refai, E.~Kou, H.~Xue, H.~Huang, W.~Xiao, G.~Wang, F.-F.
  Li, G.~Shi, et~al.
\newblock Cap-x: A framework for benchmarking and improving coding agents for
  robot manipulation.
\newblock \emph{arXiv preprint arXiv:2603.22435}, 2026.

\bibitem[Kim et~al.(2025)Kim, Yi, Choi, Ma, Goldberg, and
  Kanazawa]{kim2025pyroki}
C.~M. Kim, B.~Yi, H.~Choi, Y.~Ma, K.~Goldberg, and A.~Kanazawa.
\newblock Pyroki: A modular toolkit for robot kinematic optimization.
\newblock In \emph{IROS}, 2025.

\bibitem[Sundaralingam et~al.(2023)Sundaralingam, Hari, Fishman, Garrett, Wyk,
  Blukis, Millane, Oleynikova, Handa, Ramos, Ratliff, and Fox]{curobo_icra23}
B.~Sundaralingam, S.~K.~S. Hari, A.~Fishman, C.~R. Garrett, K.~V. Wyk,
  V.~Blukis, A.~Millane, H.~Oleynikova, A.~Handa, F.~Ramos, N.~D. Ratliff, and
  D.~Fox.
\newblock Curobo: Parallelized collision-free robot motion generation.
\newblock In \emph{{IEEE} International Conference on Robotics and Automation,
  {ICRA} 2023, London, UK, May 29 - June 2, 2023}, pages 8112--8119. {IEEE},
  2023.
\newblock \doi{10.1109/ICRA48891.2023.10160765}.
\newblock URL \url{https://doi.org/10.1109/ICRA48891.2023.10160765}.

\bibitem[Wang et~al.(2023)Wang, Zhang, Chen, Xu, Li, Liu, and
  Wang]{dexgraspnet}
R.~Wang, J.~Zhang, J.~Chen, Y.~Xu, P.~Li, T.~Liu, and H.~Wang.
\newblock Dexgraspnet: A large-scale robotic dexterous grasp dataset for
  general objects based on simulation.
\newblock In \emph{2023 IEEE International Conference on Robotics and
  Automation (ICRA)}, pages 11359--11366. IEEE, 2023.

\bibitem[Ye et~al.(2025)Ye, Wang, Yuan, Yang, Li, Zhu, Qin, Zou, and
  Wang]{ye2025dex1b}
J.~Ye, K.~Wang, C.~Yuan, R.~Yang, Y.~Li, J.~Zhu, Y.~Qin, X.~Zou, and X.~Wang.
\newblock Dex1b: Learning with 1b demonstrations for dexterous manipulation.
\newblock In \emph{RSS}, 2025.

\bibitem[Yang et~al.(2025)Yang, Liu, Li, Khaky, Shaw, and Pathak]{yang2025deep}
J.~Yang, J.~J. Liu, Y.~Li, Y.~Khaky, K.~Shaw, and D.~Pathak.
\newblock Deep reactive policy: Learning reactive manipulator motion planning
  for dynamic environments.
\newblock \emph{arXiv preprint arXiv:2509.06953}, 2025.

\bibitem[Dalal et~al.(2024)Dalal, Yang, Mendonca, Khaky, Salakhutdinov, and
  Pathak]{dalal2024neural}
M.~Dalal, J.~Yang, R.~Mendonca, Y.~Khaky, R.~Salakhutdinov, and D.~Pathak.
\newblock Neural mp: A generalist neural motion planner.
\newblock \emph{arXiv preprint arXiv:2409.05864}, 2024.

\bibitem[Yang et~al.(2025)Yang, Huang, Wu, Kanazawa, Abbeel, Sferrazza, Liu,
  Duan, and Shi]{yang2025omniretarget}
L.~Yang, X.~Huang, Z.~Wu, A.~Kanazawa, P.~Abbeel, C.~Sferrazza, C.~K. Liu,
  R.~Duan, and G.~Shi.
\newblock Omniretarget: Interaction-preserving data generation for humanoid
  whole-body loco-manipulation and scene interaction.
\newblock \emph{arXiv preprint arXiv:2509.26633}, 2025.

\bibitem[Xue et~al.(2025)Xue, Pan, Yi, Qu, and Shi]{xue2025dial}
H.~Xue, C.~Pan, Z.~Yi, G.~Qu, and G.~Shi.
\newblock Full-order sampling-based {MPC} for torque-level locomotion control
  via diffusion-style annealing.
\newblock In \emph{IEEE International Conference on Robotics and Automation
  (ICRA)}, 2025.

\bibitem[Pan et~al.(2025)Pan, Wang, Qi, Urain, Liu, Bharadhwaj, Sharma, Wu,
  Shi, Malik, and Hogan]{pan2025spider}
C.~Pan, C.~Wang, H.~Qi, J.~Urain, Z.~Liu, H.~Bharadhwaj, A.~Sharma, T.~Wu,
  G.~Shi, J.~Malik, and F.~Hogan.
\newblock Spider: Scalable physics-informed dexterous retargeting.
\newblock \emph{arXiv preprint arXiv:2511.09484}, 2025.

\bibitem[Amos and Kolter(2017)]{amos2017optnet}
B.~Amos and J.~Z. Kolter.
\newblock Optnet: Differentiable optimization as a layer in neural networks.
\newblock In \emph{International conference on machine learning}, pages
  136--145. PMLR, 2017.

\bibitem[Pineda et~al.(2022)Pineda, Fan, Monge, Venkataraman, Sodhi, Chen,
  Ortiz, DeTone, Wang, Anderson, Dong, Amos, and Mukadam]{pineda2022theseus}
L.~Pineda, T.~Fan, M.~Monge, S.~Venkataraman, P.~Sodhi, R.~T. Chen, J.~Ortiz,
  D.~DeTone, A.~Wang, S.~Anderson, J.~Dong, B.~Amos, and M.~Mukadam.
\newblock {Theseus: A Library for Differentiable Nonlinear Optimization}.
\newblock \emph{Advances in Neural Information Processing Systems}, 2022.

\bibitem[Yin et~al.(2025)Yin, Wang, Pineda, Bodduluri, Wu, Abbeel, and
  Mukadam]{yin2025geometric}
Z.-H. Yin, C.~Wang, L.~Pineda, K.~Bodduluri, T.~Wu, P.~Abbeel, and M.~Mukadam.
\newblock Geometric retargeting: A principled, ultrafast neural hand
  retargeting algorithm.
\newblock \emph{arXiv preprint arXiv:2503.07541}, 2025.

\bibitem[Zurbr{\"u}gg et~al.(2025)Zurbr{\"u}gg, Cramariuc, and
  Hutter]{zurbrugg2025graspqp}
R.~Zurbr{\"u}gg, A.~Cramariuc, and M.~Hutter.
\newblock Graspqp: Differentiable optimization of force closure for diverse and
  robust dexterous grasping.
\newblock \emph{arXiv preprint arXiv:2508.15002}, 2025.

\bibitem[Macklin(2022)]{warp2022}
M.~Macklin.
\newblock Warp: A high-performance python framework for gpu simulation and
  graphics.
\newblock \url{https://github.com/nvidia/warp}, 2022.
\newblock NVIDIA GPU Technology Conference (GTC).

\bibitem[Chen et~al.(2023)Chen, Qin, Zhou, and Su]{chen2023easyhec}
L.~Chen, Y.~Qin, X.~Zhou, and H.~Su.
\newblock Easyhec: Accurate and automatic hand-eye calibration via
  differentiable rendering and space exploration.
\newblock \emph{IEEE Robotics and Automation Letters (RA-L)}, 2023.

\bibitem[Williams et~al.(2017)Williams, Aldrich, and
  Theodorou]{williams2017model}
G.~Williams, A.~Aldrich, and E.~A. Theodorou.
\newblock Model predictive path integral control: From theory to parallel
  computation.
\newblock \emph{Journal of Guidance, Control, and Dynamics}, 40\penalty0
  (2):\penalty0 344--357, 2017.

\bibitem[Zakka et~al.(2026)Zakka, Liao, Yi, Le~Lay, Sreenath, and
  Abbeel]{zakka2026mjlab}
K.~Zakka, Q.~Liao, B.~Yi, L.~Le~Lay, K.~Sreenath, and P.~Abbeel.
\newblock mjlab: A lightweight framework for {GPU}-accelerated robot learning.
\newblock \emph{arXiv preprint arXiv:2601.22074}, 2026.

\bibitem[Diankov(2010)]{diankov2010automated}
R.~Diankov.
\newblock \emph{Automated construction of robotic manipulation programs}.
\newblock PhD thesis, Carnegie Mellon University, USA, 2010.

\bibitem[Pieper(1968)]{pieper1969kinematics}
D.~L. Pieper.
\newblock \emph{The kinematics of manipulators under computer control}.
\newblock PhD thesis, Stanford University, 1968.

\bibitem[Beeson and Ames(2015)]{beeson2015trac}
P.~Beeson and B.~Ames.
\newblock {TRAC-IK:} an open-source library for improved solving of generic
  inverse kinematics.
\newblock In \emph{15th {IEEE-RAS} International Conference on Humanoid Robots,
  Humanoids 2015, Seoul, South Korea, November 3-5, 2015}, pages 928--935.
  {IEEE}, 2015.
\newblock \doi{10.1109/HUMANOIDS.2015.7363472}.
\newblock URL \url{https://doi.org/10.1109/HUMANOIDS.2015.7363472}.

\bibitem[Zakka(2025)]{mink}
K.~Zakka.
\newblock {Mink: Python inverse kinematics based on MuJoCo}, Dec. 2025.
\newblock URL \url{https://github.com/kevinzakka/mink}.

\bibitem[Caron et~al.(2026)Caron, De~Mont-Marin, Budhiraja, Bang, Domrachev,
  Nedelchev, Du, Escande, Vaillant, Wingo, and Patapati]{pink}
S.~Caron, Y.~De~Mont-Marin, R.~Budhiraja, S.~H. Bang, I.~Domrachev,
  S.~Nedelchev, P.~Du, A.~Escande, J.~Vaillant, B.~Wingo, and S.~Patapati.
\newblock {Pink: Python inverse kinematics based on Pinocchio}, 2026.
\newblock URL \url{https://github.com/stephane-caron/pink}.

\bibitem[He et~al.(2024)He, Luo, Xiao, Zhang, Kitani, Liu, and
  Shi]{he2024learning}
T.~He, Z.~Luo, W.~Xiao, C.~Zhang, K.~Kitani, C.~Liu, and G.~Shi.
\newblock Learning human-to-humanoid real-time whole-body teleoperation.
\newblock \emph{arXiv preprint arXiv:2403.04436}, 2024.

\bibitem[Aberman et~al.(2020)Aberman, Li, Lischinski, Sorkine-Hornung,
  Cohen-Or, and Chen]{aberman2020skeleton}
K.~Aberman, P.~Li, D.~Lischinski, O.~Sorkine-Hornung, D.~Cohen-Or, and B.~Chen.
\newblock Skeleton-aware networks for deep motion retargeting.
\newblock \emph{ACM Transactions on Graphics (TOG)}, 39\penalty0 (4):\penalty0
  62--1, 2020.

\bibitem[Won and Lee(2019)]{won2019bodyshapevariation}
J.~Won and J.~Lee.
\newblock Learning body shape variation in physics-based characters.
\newblock \emph{ACM Transactions on Graphics}, 38\penalty0 (6), 2019.

\bibitem[Ryu et~al.(2021)Ryu, Kim, Lee, Park, Lee, and
  Lee]{ryu2021muscleretarget}
H.~Ryu, M.~Kim, S.~Lee, M.~S. Park, K.~Lee, and J.~Lee.
\newblock Functionality-driven musculature retargeting.
\newblock \emph{Computer Graphics Forum}, 40\penalty0 (1):\penalty0 341--356,
  2021.

\bibitem[Villegas et~al.(2021)Villegas, Ceylan, Hertzmann, Yang, and
  Saito]{villegas2021skinnedmotion}
R.~Villegas, D.~Ceylan, A.~Hertzmann, J.~Yang, and J.~Saito.
\newblock Contact-aware retargeting of skinned motion.
\newblock In \emph{Proceedings of the IEEE/CVF International Conference on
  Computer Vision}, pages 9720--9729, 2021.

\bibitem[Zhang et~al.(2023)Zhang, Gopinath, Ye, Hodgins, Turk, and
  Won]{zhang2023multicharacter}
Y.~Zhang, D.~Gopinath, Y.~Ye, J.~Hodgins, G.~Turk, and J.~Won.
\newblock Simulation and retargeting of complex multi-character interactions.
\newblock In \emph{ACM SIGGRAPH 2023 Conference Proceedings}, pages 1--11,
  2023.

\bibitem[Rajeswaran et~al.(2018)Rajeswaran, Kumar, Gupta, Vezzani, Schulman,
  Todorov, and Levine]{rajeswaran2018dapg}
A.~Rajeswaran, V.~Kumar, A.~Gupta, G.~Vezzani, J.~Schulman, E.~Todorov, and
  S.~Levine.
\newblock Learning complex dexterous manipulation with deep reinforcement
  learning and demonstrations.
\newblock In \emph{Proceedings of Robotics: Science and Systems (RSS)}, 2018.

\bibitem[Kumar and Todorov(2015)]{kumar2015haptix}
V.~Kumar and E.~Todorov.
\newblock Mujoco haptix: A virtual reality system for hand manipulation.
\newblock In \emph{2015 IEEE-RAS 15th International Conference on Humanoid
  Robots (Humanoids)}, pages 657--663, 2015.

\bibitem[Humberston and Pai(2015)]{humbertson2015handson}
B.~Humberston and D.~K. Pai.
\newblock Hands on: Interactive animation of precision manipulation and
  contact.
\newblock In \emph{Proceedings of the 14th ACM SIGGRAPH / Eurographics
  Symposium on Computer Animation}, page 63–72, 2015.

\bibitem[Qin et~al.(2022)Qin, Su, and Wang]{qin2022from}
Y.~Qin, H.~Su, and X.~Wang.
\newblock From one hand to multiple hands: Imitation learning for dexterous
  manipulation from single-camera teleoperation.
\newblock \emph{IEEE Robotics and Automation Letters}, 7:\penalty0
  10873--10881, 2022.

\bibitem[Antotsiou et~al.(2018)Antotsiou, Garcia-Hernando, and
  K.]{antotsiou2018taskretargeting}
D.~Antotsiou, G.~Garcia-Hernando, and K.~T. K.
\newblock Task-oriented hand motion retargeting for dexterous manipulation
  imitation.
\newblock In \emph{The European Conference on Computer Vision (ECCV)
  Workshops}, 2018.

\bibitem[Dasari et~al.(2023)Dasari, Gupta, and Kumar]{dasari2023pgdm}
S.~Dasari, A.~Gupta, and V.~Kumar.
\newblock Learning dexterous manipulation from exemplar object trajectories and
  pre-grasps.
\newblock In \emph{IEEE International Conference on Robotics and Automation},
  2023.

\bibitem[Kry and Pai(2006)]{kry2006interactioncapture}
P.~G. Kry and D.~K. Pai.
\newblock Interaction capture and synthesis.
\newblock \emph{ACM Transactions on Graphics}, 25\penalty0 (3):\penalty0
  872–880, 2006.

\bibitem[Handa et~al.(2019)Handa, Wyk, Yang, Liang, Chao, Wan, Birchfield,
  Ratliff, and Fox]{Handa2019DexPilotVT}
A.~Handa, K.~V. Wyk, W.~Yang, J.~Liang, Y.~W. Chao, Q.~Wan, S.~Birchfield,
  N.~D. Ratliff, and D.~Fox.
\newblock Dexpilot: Vision-based teleoperation of dexterous robotic hand-arm
  system.
\newblock \emph{2020 IEEE International Conference on Robotics and Automation
  (ICRA)}, pages 9164--9170, 2019.

\bibitem[Sivakumar et~al.(2022)Sivakumar, Shaw, and
  Pathak]{sivakumar2022telekinesis}
A.~Sivakumar, K.~Shaw, and D.~Pathak.
\newblock Robotic telekinesis: Learning a robotic hand imitator by watching
  humans on youtube.
\newblock In \emph{Proceedings of Robotics: Science and Systems (RSS)}, 2022.

\bibitem[Wu et~al.(2022)Wu, Wang, and Wang]{wu2022ILAD}
Y.~H. Wu, J.~Wang, and X.~Wang.
\newblock Learning generalizable dexterous manipulation from human grasp
  affordance.
\newblock In \emph{Conference on Robot Learning (CORL)}, 2022.

\bibitem[Li et~al.(2025)Li, Li, Liu, Li, and Huang]{li2025maniptrans}
K.~Li, P.~Li, T.~Liu, Y.~Li, and S.~Huang.
\newblock {ManipTrans}: Efficient dexterous bimanual manipulation transfer via
  residual learning.
\newblock In \emph{Proceedings of the IEEE/CVF Conference on Computer Vision
  and Pattern Recognition (CVPR)}, 2025.

\bibitem[Mandi et~al.(2025)Mandi, Hou, Fox, Narang, Mandlekar, and
  Song]{mandi2025dexmachina}
Z.~Mandi, Y.~Hou, D.~Fox, Y.~Narang, A.~Mandlekar, and S.~Song.
\newblock {DexMachina}: Functional retargeting for bimanual dexterous
  manipulation.
\newblock \emph{arXiv preprint arXiv:2505.24853}, 2025.

\bibitem[Miller and Allen(2004)]{miller2004graspit}
A.~T. Miller and P.~K. Allen.
\newblock Graspit! a versatile simulator for robotic grasping.
\newblock In \emph{IEEE Robotics \& Automation Magazine}, 2004.

\bibitem[Zhu et~al.(2021)Zhu, Wu, Lin, and Sun]{human-like-grasp}
T.~Zhu, R.~Wu, X.~Lin, and Y.~Sun.
\newblock Toward human-like grasp: Dexterous grasping via semantic
  representation of object-hand.
\newblock In \emph{Proceedings of the IEEE/CVF International Conference on
  Computer Vision}, 2021.

\bibitem[Taheri et~al.(2020)Taheri, Ghorbani, Black, and
  Tzionas]{taheri2020grab}
O.~Taheri, N.~Ghorbani, M.~J. Black, and D.~Tzionas.
\newblock Grab: A dataset of whole-body human grasping of objects.
\newblock In \emph{European conference on computer vision}, pages 581--600.
  Springer, 2020.

\bibitem[Liu et~al.(2024)Liu, Yang, Wang, Wu, Wang, Yao, Schwertfeger, Yang,
  Wang, Yu, et~al.]{liu2024realdex}
Y.~Liu, Y.~Yang, Y.~Wang, X.~Wu, J.~Wang, Y.~Yao, S.~Schwertfeger, S.~Yang,
  W.~Wang, J.~Yu, et~al.
\newblock Realdex: towards human-like grasping for robotic dexterous hand.
\newblock In \emph{Proceedings of the Thirty-Third International Joint
  Conference on Artificial Intelligence}, pages 6859--6867, 2024.

\bibitem[Liu et~al.(2021)Liu, Liu, Jiao, Zhu, and Zhu]{liu2021dfc}
T.~Liu, Z.~Liu, Z.~Jiao, Y.~Zhu, and S.-C. Zhu.
\newblock Synthesizing diverse and physically stable grasps with arbitrary hand
  structures using differentiable force closure estimator.
\newblock \emph{IEEE Robotics and Automation Letters}, 7\penalty0 (1):\penalty0
  470--477, 2021.

\bibitem[Li et~al.(2023)Li, Liu, Li, Geng, Zhu, Yang, and Huang]{gendexgrasp}
P.~Li, T.~Liu, Y.~Li, Y.~Geng, Y.~Zhu, Y.~Yang, and S.~Huang.
\newblock Gendexgrasp: Generalizable dexterous grasping.
\newblock In \emph{2023 IEEE International Conference on Robotics and
  Automation (ICRA)}, pages 8068--8074. IEEE, 2023.

\bibitem[Shao et~al.(2020)Shao, Ferreira, Jorda, Nambiar, Luo, Solowjow, Ojea,
  Khatib, and Bohg]{unigrasp}
L.~Shao, F.~Ferreira, M.~Jorda, V.~Nambiar, J.~Luo, E.~Solowjow, J.~A. Ojea,
  O.~Khatib, and J.~Bohg.
\newblock Unigrasp: Learning a unified model to grasp with multifingered
  robotic hands.
\newblock \emph{IEEE Robotics and Automation Letters}, 5\penalty0 (2):\penalty0
  2286--2293, 2020.

\bibitem[Liu et~al.(2023)Liu, Zhou, Yang, Gupta, and Wang]{contactgen}
S.~Liu, Y.~Zhou, J.~Yang, S.~Gupta, and S.~Wang.
\newblock Contactgen: Generative contact modeling for grasp generation.
\newblock In \emph{Proceedings of the IEEE/CVF International Conference on
  Computer Vision}, pages 20609--20620, 2023.

\bibitem[Xu et~al.(2023)Xu, Wan, Zhang, Liu, Shan, Shen, Wang, Geng, Weng,
  Chen, et~al.]{unidexgrasp}
Y.~Xu, W.~Wan, J.~Zhang, H.~Liu, Z.~Shan, H.~Shen, R.~Wang, H.~Geng, Y.~Weng,
  J.~Chen, et~al.
\newblock Unidexgrasp: Universal robotic dexterous grasping via learning
  diverse proposal generation and goal-conditioned policy.
\newblock In \emph{Proceedings of the IEEE/CVF Conference on Computer Vision
  and Pattern Recognition}, pages 4737--4746, 2023.

\bibitem[Wan et~al.(2023)Wan, Geng, Liu, Shan, Yang, Yi, and
  Wang]{unidexgrasp++}
W.~Wan, H.~Geng, Y.~Liu, Z.~Shan, Y.~Yang, L.~Yi, and H.~Wang.
\newblock Unidexgrasp++: Improving dexterous grasping policy learning via
  geometry-aware curriculum and iterative generalist-specialist learning.
\newblock In \emph{Proceedings of the IEEE/CVF International Conference on
  Computer Vision}, pages 3891--3902, 2023.

\bibitem[Wu et~al.(2024)Wu, Wu, Zhang, Gan, and Dong]{graspgf}
T.~Wu, M.~Wu, J.~Zhang, Y.~Gan, and H.~Dong.
\newblock Learning score-based grasping primitive for human-assisting dexterous
  grasping.
\newblock \emph{Advances in Neural Information Processing Systems}, 2024.

\bibitem[Qin et~al.(2023)Qin, Huang, Yin, Su, and Wang]{dexpoint}
Y.~Qin, B.~Huang, Z.-H. Yin, H.~Su, and X.~Wang.
\newblock Dexpoint: Generalizable point cloud reinforcement learning for
  sim-to-real dexterous manipulation.
\newblock In \emph{Conference on Robot Learning}, 2023.

\bibitem[Chen et~al.(2025)Chen, Ke, and Wang]{chen2025bodex}
J.~Chen, Y.~Ke, and H.~Wang.
\newblock Bodex: Scalable and efficient robotic dexterous grasp synthesis using
  bilevel optimization.
\newblock In \emph{2025 IEEE International Conference on Robotics and
  Automation (ICRA)}, pages 01--08. IEEE, 2025.

\bibitem[Li et~al.(2024)Li, Mao, Deng, Meng, Fan, Wang, Osamu, Tan, Wang, and
  Deng]{li2024multigraspllm}
H.~Li, W.~Mao, W.~Deng, C.~Meng, H.~Fan, T.~Wang, Y.~Osamu, P.~Tan, H.~Wang,
  and X.~Deng.
\newblock Multi-graspllm: A multimodal llm for multi-hand semantic guided grasp
  generation.
\newblock \emph{arXiv preprint arXiv:2412.08468}, 2024.

\bibitem[Schulman et~al.(2014)Schulman, Duan, Ho, Lee, Awwal, Bradlow, Pan,
  Patil, Goldberg, and Abbeel]{schulman2014motion}
J.~Schulman, Y.~Duan, J.~Ho, A.~X. Lee, I.~Awwal, H.~Bradlow, J.~Pan, S.~Patil,
  K.~Goldberg, and P.~Abbeel.
\newblock Motion planning with sequential convex optimization and convex
  collision checking.
\newblock \emph{The International Journal of Robotics Research}, 33\penalty0
  (9):\penalty0 1251--1270, 2014.
\newblock \doi{10.1177/0278364914528132}.
\newblock URL \url{https://doi.org/10.1177/0278364914528132}.

\bibitem[Chitta et~al.(2012)Chitta, Sucan, and Cousins]{chitta2012moveit}
S.~Chitta, I.~Sucan, and S.~Cousins.
\newblock Moveit![ros topics].
\newblock \emph{IEEE robotics \& automation magazine}, 19\penalty0
  (1):\penalty0 18--19, 2012.

\bibitem[Tedrake and {the Drake Development Team}(2019)]{drake}
R.~Tedrake and {the Drake Development Team}.
\newblock Drake: Model-based design and verification for robotics.
\newblock \url{https://drake.mit.edu}, 2019.

\bibitem[Mastalli et~al.(2020)Mastalli, Budhiraja, Merkt, Saurel, Hammoud,
  Naveau, Carpentier, Righetti, Vijayakumar, and
  Mansard]{mastalli2020crocoddyl}
C.~Mastalli, R.~Budhiraja, W.~Merkt, G.~Saurel, B.~Hammoud, M.~Naveau,
  J.~Carpentier, L.~Righetti, S.~Vijayakumar, and N.~Mansard.
\newblock Crocoddyl: An efficient and versatile framework for multi-contact
  optimal control.
\newblock In \emph{IEEE International Conference on Robotics and Automation
  (ICRA)}, 2020.

\bibitem[Thomason et~al.(2024)Thomason, Kingston, and
  Kavraki]{thomason2024vamp}
W.~Thomason, Z.~Kingston, and L.~E. Kavraki.
\newblock Motions in microseconds via vectorized sampling-based planning.
\newblock In \emph{IEEE International Conference on Robotics and Automation
  (ICRA)}, 2024.

\bibitem[Sundaralingam et~al.(2026)Sundaralingam, Murali, and
  Birchfield]{sundaralingam2026curobov2}
B.~Sundaralingam, A.~Murali, and S.~Birchfield.
\newblock curobov2: Dynamics-aware motion generation with depth-fused distance
  fields for high-dof robots.
\newblock \emph{arXiv preprint arXiv:2603.05493}, 2026.

\bibitem[Howell et~al.(2022)Howell, Gileadi, Tunyasuvunakool, Zakka, Erez, and
  Tassa]{howell2022predictive}
T.~Howell, N.~Gileadi, S.~Tunyasuvunakool, K.~Zakka, T.~Erez, and Y.~Tassa.
\newblock Predictive sampling: Real-time behaviour synthesis with {MuJoCo}.
\newblock \emph{arXiv preprint arXiv:2212.00541}, 2022.

\bibitem[Hansen(2016)]{hansen2016cma}
N.~Hansen.
\newblock The {CMA} evolution strategy: A tutorial.
\newblock \emph{arXiv preprint arXiv:1604.00772}, 2016.

\bibitem[{Google DeepMind and NVIDIA}(2025)]{mujocowarp2025}
{Google DeepMind and NVIDIA}.
\newblock {MuJoCo Warp}: {GPU}-optimized version of the {MuJoCo} physics
  simulator.
\newblock \url{https://github.com/google-deepmind/mujoco_warp}, 2025.

\bibitem[Mahmood et~al.(2019)Mahmood, Ghorbani, Troje, Pons-Moll, and
  Black]{mahmood2019amass}
N.~Mahmood, N.~Ghorbani, N.~F. Troje, G.~Pons-Moll, and M.~J. Black.
\newblock Amass: Archive of motion capture as surface shapes.
\newblock In \emph{Proceedings of the IEEE/CVF international conference on
  computer vision}, pages 5442--5451, 2019.

\bibitem[Makoviychuk et~al.(2021)Makoviychuk, Wawrzyniak, Guo, Lu, Storey,
  Macklin, Hoeller, Rudin, Allshire, Handa, and State]{makoviychuk2021isaac}
V.~Makoviychuk, L.~Wawrzyniak, Y.~Guo, M.~Lu, K.~Storey, M.~Macklin,
  D.~Hoeller, N.~Rudin, A.~Allshire, A.~Handa, and G.~State.
\newblock Isaac gym: High performance {GPU} based physics simulation for robot
  learning.
\newblock In \emph{NeurIPS Datasets and Benchmarks Track}, 2021.

\bibitem[Zakka et~al.(2025)Zakka, Tabanpour, Liao, Haiderbhai, Holt, Luo,
  Allshire, Frey, Sreenath, Kahrs, Sferrazza, Tassa, and
  Abbeel]{zakka2025mujoco}
K.~Zakka, B.~Tabanpour, Q.~Liao, M.~Haiderbhai, S.~Holt, J.~Y. Luo,
  A.~Allshire, E.~Frey, K.~Sreenath, L.~A. Kahrs, C.~Sferrazza, Y.~Tassa, and
  P.~Abbeel.
\newblock {MuJoCo Playground}.
\newblock \emph{arXiv preprint arXiv:2502.08844}, 2025.

\bibitem[Todorov et~al.(2012)Todorov, Erez, and Tassa]{todorov2012mujoco}
E.~Todorov, T.~Erez, and Y.~Tassa.
\newblock {MuJoCo}: A physics engine for model-based control.
\newblock In \emph{IEEE/RSJ International Conference on Intelligent Robots and
  Systems (IROS)}, pages 5026--5033, 2012.

\bibitem[Zhan et~al.(2024)Zhan, Yang, Zhao, Mao, Xu, Lin, Li, and
  Lu]{zhan2024oakink2}
X.~Zhan, L.~Yang, Y.~Zhao, K.~Mao, H.~Xu, Z.~Lin, K.~Li, and C.~Lu.
\newblock {OAKINK2}: A dataset of bimanual hands-object manipulation in complex
  task completion.
\newblock In \emph{Proceedings of the IEEE/CVF Conference on Computer Vision
  and Pattern Recognition (CVPR)}, 2024.

\bibitem[Fu et~al.(2025)Fu, Zhang, Jiang, Fu, Funk, Ritchie, and
  Sridhar]{fu2025gigahands}
R.~Fu, D.~Zhang, A.~Jiang, W.~Fu, A.~Funk, D.~Ritchie, and S.~Sridhar.
\newblock {GigaHands}: A massive annotated dataset of bimanual hand activities.
\newblock In \emph{Proceedings of the Computer Vision and Pattern Recognition
  Conference (CVPR)}, 2025.

\bibitem[Lu et~al.(2024)Lu, Kang, Li, Liu, Yang, Huang, and Hua]{lu2024ugg}
J.~Lu, H.~Kang, H.~Li, B.~Liu, Y.~Yang, Q.~Huang, and G.~Hua.
\newblock Ugg: Unified generative grasping.
\newblock In \emph{ECCV}, 2024.

\bibitem[Chamzas et~al.(2022)Chamzas, Quintero-Pe{\~n}a, Kingston, Orthey,
  Rakita, Gleicher, Toussaint, and Kavraki]{chamzas2022motionbenchmaker}
C.~Chamzas, C.~Quintero-Pe{\~n}a, Z.~Kingston, A.~Orthey, D.~Rakita,
  M.~Gleicher, M.~Toussaint, and L.~E. Kavraki.
\newblock {MotionBenchMaker}: A tool to generate and benchmark motion planning
  datasets.
\newblock \emph{IEEE Robotics and Automation Letters}, 7\penalty0 (2):\penalty0
  882--889, 2022.

\bibitem[Fishman et~al.(2022)Fishman, Murali, Eppner, Peele, Boots, and
  Fox]{fishman2022motion}
A.~Fishman, A.~Murali, C.~Eppner, B.~Peele, B.~Boots, and D.~Fox.
\newblock Motion policy networks.
\newblock In \emph{Conference on Robot Learning (CoRL)}, 2022.

\bibitem[Chao et~al.(2021)Chao, Yang, Xiang, Molchanov, Handa, Tremblay,
  Narang, Van~Wyk, Iqbal, Birchfield, et~al.]{chao2021dexycb}
Y.-W. Chao, W.~Yang, Y.~Xiang, P.~Molchanov, A.~Handa, J.~Tremblay, Y.~S.
  Narang, K.~Van~Wyk, U.~Iqbal, S.~Birchfield, et~al.
\newblock Dexycb: A benchmark for capturing hand grasping of objects.
\newblock In \emph{Proceedings of the IEEE/CVF conference on computer vision
  and pattern recognition}, pages 9044--9053, 2021.

\bibitem[Shin et~al.(2024)Shin, Kim, Halilaj, and Black]{shin2024wham}
S.~Shin, J.~Kim, E.~Halilaj, and M.~J. Black.
\newblock {WHAM}: Reconstructing world-grounded humans with accurate {3D}
  motion.
\newblock In \emph{Proceedings of the IEEE/CVF Conference on Computer Vision
  and Pattern Recognition (CVPR)}, 2024.

\bibitem[Romero et~al.(2017)Romero, Tzionas, and Black]{romero2017embodied}
J.~Romero, D.~Tzionas, and M.~J. Black.
\newblock Embodied hands: Modeling and capturing hands and bodies together.
\newblock \emph{ACM Transactions on Graphics (Proc. SIGGRAPH Asia)},
  36\penalty0 (6), 2017.

\bibitem[Tao et~al.(2024)Tao, Xiang, Shukla, Qin, Hinrichsen, Yuan, Bao, Lin,
  Liu, Chan, et~al.]{tao2024maniskill3}
S.~Tao, F.~Xiang, A.~Shukla, Y.~Qin, X.~Hinrichsen, X.~Yuan, C.~Bao, X.~Lin,
  Y.~Liu, T.-k. Chan, et~al.
\newblock Maniskill3: Gpu parallelized robotics simulation and rendering for
  generalizable embodied ai.
\newblock \emph{arXiv preprint arXiv:2410.00425}, 2024.

\bibitem[Schulman et~al.(2017)Schulman, Wolski, Dhariwal, Radford, and
  Klimov]{schulman2017proximal}
J.~Schulman, F.~Wolski, P.~Dhariwal, A.~Radford, and O.~Klimov.
\newblock Proximal policy optimization algorithms.
\newblock \emph{arXiv preprint arXiv:1707.06347}, 2017.

\bibitem[Carpentier et~al.(2019)Carpentier, Saurel, Buondonno, Mirabel,
  Lamiraux, Stasse, and Mansard]{carpentier2019pinocchio}
J.~Carpentier, G.~Saurel, G.~Buondonno, J.~Mirabel, F.~Lamiraux, O.~Stasse, and
  N.~Mansard.
\newblock The pinocchio c++ library: A fast and flexible implementation of
  rigid body dynamics algorithms and their analytical derivatives.
\newblock In \emph{2019 IEEE/SICE International Symposium on System Integration
  (SII)}, pages 614--619. IEEE, 2019.

\end{thebibliography}

\clearpage
\appendix
\section*{Appendix}
In this supplementary material, we first detail the experimental settings, including the hardware, per-task definitions and metrics, and baselines. Second, we provide additional experiments that complement the paper. Third, we describe the implementation details, including the list of built-in cost terms and the algorithm details. Finally, we present additional qualitative results. Please see our \textbf{video} for an overview of our work and additional visualizations.

\section{Experimental Details}

\subsection{Hardware and Profiling}
\vspace{-5pt}

Our benchmark platform consists of a single NVIDIA RTX 4090 GPU paired with an Intel Xeon Gold 5120 CPU @ $2.20$\,GHz. Unless otherwise noted, all reported timings are averaged over $5$ runs after warmup, which ensures that GPU kernels are compiled and cached before measurement. Unless stated otherwise, time refers to the total wall-clock solve time for the full batch.

\subsection{Task Definitions}
\vspace{-5pt}

We evaluate RoboKit on six kinematic tasks, one physics-based task, and RL training. For each task we describe the dataset and embodiment, the optimization objective, and the evaluation protocol.

\noindent
\textbf{Inverse Kinematics.} Following the cuRobo~\cite{curobo_icra23} benchmarks, we solve IK on a fixed-base 7-DoF Franka Panda arm. The objective combines the pose-matching cost $c_{\mathrm{pose}}$ for end-effector alignment with the joint-limit cost $c_{\mathrm{limit}}$. We report solve time, success rate (position error $<5$\,mm and orientation error $<0.05$\,rad), and $98$th-percentile position (mm) and orientation (rad) errors over successful instances across batch sizes from $1$ to $2000$.

\noindent
\textbf{Trajectory Optimization.} We generate collision-free reaching trajectories for the Franka arm in cluttered scenes from MotionBenchMaker~\cite{chamzas2022motionbenchmaker} and MPiNets~\cite{fishman2022motion}, following the cuRobo~\cite{curobo_icra23} benchmark. The objective includes the goal pose-matching cost $c_{\mathrm{pose}}$, smoothness cost $c_{\mathrm{smooth}}$, joint-limit cost $c_{\mathrm{limit}}$, and collision-avoidance cost $c_{\mathrm{coll}}$. A trajectory is counted as successful if it reaches the goal (position error $\leq5$\,mm and $\sin(\theta/2)\leq0.05$, where $\theta$ is the orientation error) and satisfies world/self-collision, joint-limit, and dynamic-limit checks on a $25$\,ms interpolation, using the original benchmark's $0.2$\,rad joint-limit tolerance. We report the mean planning time (Plan), its $98$th percentile (P98), the executed motion time after time-optimal parameterization (Motion), the total joint-space path length (Path, in rad), maximum absolute joint jerk per trajectory (Max Jerk, rad/s$^3$), mean absolute joint velocity over frames and joints (Mean Vel, rad/s), and maximum absolute joint acceleration per trajectory (Max Acc, rad/s$^2$). Plan and P98 use external CUDA-synchronized wall-clock solve time; all these metrics are aggregated over successful plans.

\noindent
\textbf{Hand Motion Retargeting.} Following Dex-Retargeting~\cite{qin2023anyteleop}, we retarget human hand motion to the Shadow robot hand and evaluate online, one frame at a time, on 501 right-hand captures from DexYCB~\cite{chao2021dexycb}. The objective uses a keypoint position-alignment cost and a vector-alignment cost together with joint-limit constraints, and a fixed scale factor accounts for morphological differences. We report per-frame solve time, wrist-relative fingertip position error (Pos., mm), pinch and open success (\%), and mean joint acceleration (Acc., rad/s$^2$). For the distilled-policy comparison against GeoRT~\cite{yin2025geometric} (Tab.~\ref{tab:geort}) we additionally report fingertip position error (Tip), its Procrustes-aligned variant (PA-Tip), and the 95th-percentile tip error (P95), all in mm.

\noindent
\textbf{Human Motion Retargeting.} We retarget full-body human motion from the SFU subset of AMASS~\cite{mahmood2019amass} ($44$ sequences) to the Unitree G1 humanoid, following GMR~\cite{joao2025gmr}. The objective applies pose-matching costs on key body links (hands, feet, and head) together with joint-limit constraints. We report per-frame solve time, root-relative limb position error (MPJPE, cm), world-frame limb position error after one similarity alignment per $100$-frame segment (WA-MPJPE, cm), as in WHAM~\cite{shin2024wham}, and mean joint acceleration (Acc., rad/s$^2$).

\noindent
\textbf{Grasp Synthesis.} We synthesize dexterous grasps for the MANO~\cite{romero2017embodied} hand on objects from the test split of DexGraspNet~\cite{dexgraspnet}, following Dex1B~\cite{ye2025dex1b}. The objective combines the collision-avoidance cost $c_{\mathrm{coll}}$, force-closure cost $c_{\mathrm{fc}}$, a contact cost that encourages finger contact with the object surface, and the joint-limit cost $c_{\mathrm{limit}}$. We synthesize $4096$ grasps ($32$ on each of $128$ test scenes) with $24$ seeds per grasp. Success requires pre-grasp penetration at most $5$\,mm and passing the ManiSkill~\cite{tao2024maniskill3} grasp test under external forces, using Dex1B's force threshold; we also report the $\epsilon$-metric grasp quality (Q1) and the penetration depth (mm).

\noindent
\textbf{Rendering-Based Calibration.} Following the EasyHEC~\cite{chen2023easyhec} benchmark, we calibrate the hand-eye transform of a Panda arm by matching rendered robot masks to observed silhouettes through our coarse-to-fine rendering pipeline (Sec.~\ref{subsec:accel}). We evaluate the realistic setting with a $35$\,cm / $45^\circ$ initial error and added sensor and calibration noise, and report calibration time (s), translation error (cm), and rotation error ($^\circ$).

\noindent
\textbf{Physics-Based Retargeting.} Following SPIDER's~\cite{pan2025spider} MuJoCo comparison, we track the human hand--object motion of $7$ bimanual OakInk tasks and $3$ GigaHands clips with the Ability, Allegro, Inspire, Schunk and XHand hands, five seeds each, using SPIDER's released kinematic retargets, scenes and configurations. Both methods run sampling-based MPC with the physics engine in the loop; we score $218$ of the $248$ released trials (the other $30$ fail SPIDER's own export check). We report the object's ADD-AUC (area under the curve of the object-position error over $0$--$10$\,cm thresholds, as SPIDER), success, and the object's mean position (cm) and rotation ($^\circ$) errors. SPIDER's accuracy is that of its released rollouts.

\noindent
\textbf{RL Training.} We train PPO~\cite{schulman2017proximal} policies on mjlab's~\cite{zakka2026mjlab} tasks with $4096$ parallel environments: Go1 and G1 velocity tracking on flat and rough terrain, G1 motion tracking, and Yam cube lifting. RoboKit runs mjlab's rewards, terminations, commands, curricula, randomization and observations on its scenes, checked against mjlab on identical states. One evaluator scores both methods' policies: linear-velocity tracking, motion-tracking success, or time at the goal. We report wall-clock minutes to reach $95\%$ and $100\%$ of mjlab's mean final score, median over five seeds.

\subsection{Baselines}
\vspace{-5pt}

We compare RoboKit against optimization-based and learned baselines, depending on the task.

\noindent
\textbf{PyRoKi}~\cite{kim2025pyroki} is a recent kinematics optimization library built on JAX that supports batched optimization with automatic differentiation. It is our primary baseline for inverse kinematics and hand motion retargeting.

\noindent
\textbf{cuRobo}~\cite{curobo_icra23} is a CUDA-accelerated motion-generation framework that uses L-BFGS with particle-based seeding. We compare against it on the batched IK benchmark (Tab.~\ref{tab:metric_curobo}) and on trajectory optimization.

\noindent
\textbf{Dex-Retargeting}~\cite{qin2023anyteleop} is a widely used library for dexterous hand retargeting, offering a position-based mode that matches joint positions and vector-based modes that align palm-to-fingertip vectors; we compare against its position and DexPilot modes. It does not support batched optimization, so we report single-instance results.

\noindent
\textbf{GMR}~\cite{joao2025gmr} retargets human motion to humanoid robots using pose-matching objectives on keypoints subject to joint limits. It is our baseline for human motion retargeting and also runs single-instance.

\noindent
\textbf{Dex1B}~\cite{ye2025dex1b} is a large-scale grasp-synthesis benchmark that generates dexterous grasps via optimization with force-closure and contact objectives; we adopt its evaluation protocol and success criteria. We additionally compare our distilled grasping policy against the learned baselines Dex1B and UGG~\cite{lu2024ugg}.

\noindent
\textbf{GeoRT}~\cite{yin2025geometric} is a learned geometric hand-retargeting method, used as a learned baseline for the distilled hand-retargeting policy.

\noindent
\textbf{EasyHEC}~\cite{chen2023easyhec} is a rendering-based hand-eye calibration method and serves as our baseline for rendering-based calibration.

\noindent
\textbf{SPIDER}~\cite{pan2025spider} retargets human hand--object motion to dexterous hands with sampling-based MPC in MuJoCo Warp; it is our baseline for physics-based retargeting, run from its released code and configurations.

\noindent
\textbf{DIAL-MPC}~\cite{xue2025dial} is a sampling-based MPC with diffusion-style annealing for torque-level legged locomotion; it is our baseline for Go2 locomotion (Appendix~\ref{sec:dial}), run from its released code.

\noindent
\textbf{mjlab}~\cite{zakka2026mjlab} is a GPU RL framework with Isaac Lab's manager-based API on MuJoCo Warp; it is our baseline for RL training (Tab.~\ref{tab:rl_mjlab}), run from its released tasks.

\noindent
\textbf{Pinocchio}~\cite{carpentier2019pinocchio} and \textbf{Pink}~\cite{pink} are widely used CPU-based Python IK libraries, which we use for the single-instance IK comparison (Tab.~\ref{tab:single_instance_ik}).

\section{Additional Experiments}

\subsection{Batch Scaling}
\label{sec:batch_scaling}

We evaluate batch scaling for IK, motion planning, hand and humanoid retargeting, and grasp synthesis. We additionally report mean$\pm$std across $5$ runs, together with IK success rates and solution errors.

\begin{table}[H]
    \tablestyle
    \caption{\textbf{Batched IK runtime and accuracy} on the 7-DoF Franka Panda.}
    \label{tab:metric_curobo}
    \metricstable{llcccc}{6}{%
        \toprule
        Batch & Method & \makecell{Time\\(ms)} & \makecell{SR\\(\%)} & \makecell{Pos. Err\\(mm)} & \makecell{Ori. Err\\(rad)} \\
        \midrule
        1 & PyRoKi~\cite{kim2025pyroki} & $9.08\pm0.72$ & $100.00\pm0.00$ & $(1.74\pm0.46)\times10^{-4}$ & $(1.88\pm1.13)\times10^{-7}$ \\
        1 & cuRobo~\cite{curobo_icra23} & $7.43\pm0.10$ & $100.00\pm0.00$ & $(2.13\pm0.94)\times10^{-3}$ & $(8.14\pm4.33)\times10^{-6}$ \\
        \rowcolor{myblue}
        1 & Ours (opt) & $0.65\pm0.02$ & $100.00\pm0.00$ & $(1.23\pm0.45)\times10^{-4}$ & $(1.68\pm1.22)\times10^{-7}$ \\
        \addlinespace[2pt]
        10 & PyRoKi~\cite{kim2025pyroki} & $14.52\pm0.70$ & $100.00\pm0.00$ & $(3.80\pm7.52)\times10^{-2}$ & $(4.86\pm9.66)\times10^{-5}$ \\
        10 & cuRobo~\cite{curobo_icra23} & $6.73\pm0.02$ & $100.00\pm0.00$ & $(3.16\pm1.52)\times10^{-3}$ & $(1.17\pm0.18)\times10^{-5}$ \\
        \rowcolor{myblue}
        10 & Ours (opt) & $0.61\pm0.02$ & $100.00\pm0.00$ & $(1.15\pm2.26)\times10^{-2}$ & $(1.90\pm3.74)\times10^{-5}$ \\
        \addlinespace[2pt]
        100 & PyRoKi~\cite{kim2025pyroki} & $16.04\pm0.27$ & $100.00\pm0.00$ & $(1.91\pm2.46)\times10^{-2}$ & $(2.13\pm3.33)\times10^{-5}$ \\
        100 & cuRobo~\cite{curobo_icra23} & $8.36\pm0.05$ & $99.80\pm0.40$ & $(4.92\pm1.78)\times10^{-3}$ & $(1.39\pm0.09)\times10^{-5}$ \\
        \rowcolor{myblue}
        100 & Ours (opt) & $0.81\pm0.01$ & $100.00\pm0.00$ & $(1.08\pm0.71)\times10^{-3}$ & $(1.20\pm0.82)\times10^{-6}$ \\
        \addlinespace[2pt]
        1000 & PyRoKi~\cite{kim2025pyroki} & $48.19\pm1.30$ & $99.88\pm0.10$ & $(1.47\pm0.83)\times10^{-3}$ & $(1.01\pm0.41)\times10^{-6}$ \\
        1000 & cuRobo~\cite{curobo_icra23} & $28.48\pm0.70$ & $99.94\pm0.05$ & $(4.39\pm0.24)\times10^{-3}$ & $(1.40\pm0.01)\times10^{-5}$ \\
        \rowcolor{myblue}
        1000 & Ours (opt) & $3.41\pm0.02$ & $99.94\pm0.05$ & $(4.04\pm1.53)\times10^{-4}$ & $(4.15\pm0.32)\times10^{-7}$ \\
        \addlinespace[2pt]
        2000 & PyRoKi~\cite{kim2025pyroki} & $86.20\pm0.73$ & $99.89\pm0.07$ & $(1.57\pm0.77)\times10^{-3}$ & $(1.05\pm0.45)\times10^{-6}$ \\
        2000 & cuRobo~\cite{curobo_icra23} & $50.27\pm0.06$ & $99.96\pm0.04$ & $(4.29\pm0.20)\times10^{-3}$ & $(1.39\pm0.00)\times10^{-5}$ \\
        \rowcolor{myblue}
        2000 & Ours (opt) & $7.12\pm0.02$ & $99.91\pm0.04$ & $(3.28\pm0.30)\times10^{-4}$ & $(4.03\pm0.25)\times10^{-7}$ \\
        \bottomrule

    }
    \par\vspace{4pt}
\end{table}

\nopagebreak[4]
\noindent
\textbf{Batched IK comparison with cuRobo.}
Table~\ref{tab:metric_curobo} extends the main-paper IK comparison with cuRobo~\cite{curobo_icra23}. RoboKit is $7.1$--$11.5\times$ faster than cuRobo, with comparable success rates and generally lower errors. Time covers the full batch; success requires position error below $5$\,mm and orientation error below $0.05$\,rad. Reported errors are per-run $98$th percentiles over successful instances.

\begin{table}[H]
    \tablestyle
    \caption{\textbf{Batch latency} in milliseconds (mean$\pm$population std across $5$ runs).}
    \label{tab:batch_scaling}
    \metricstable{llcccc}{6}{%
        \toprule
        Task & Method & \makecell{Batch\\1} & \makecell{Batch\\10} & \makecell{Batch\\100} & \makecell{Batch\\1000} \\
        \midrule
        Motion & cuRobo & $48.61\pm1.06$ & $57.69\pm1.19$ & $335.66\pm1.99$ & n.m. \\
        \rowcolor{myblue}
        Motion & Ours (opt) & $13.27\pm0.57$ & $30.54\pm0.54$ & $230.90\pm0.04$ & n.m. \\
        \addlinespace[3pt]
        Hand & Dex-Ret. (pos) & $10.27\pm0.12$ & -- & -- & -- \\
        Hand & DexPilot & $17.06\pm0.33$ & -- & -- & -- \\
        Hand & PyRoKi & $17.72\pm1.99$ & $248.46\pm1.11$ & $648.33\pm3.47$ & $1106.27\pm132.51$ \\
        \rowcolor{myblue}
        Hand & Ours (opt) & $1.37\pm0.02$ & $1.37\pm0.01$ & $1.64\pm0.12$ & $3.02\pm0.09$ \\
        \addlinespace[3pt]
        Humanoid & GMR & $15.478\pm0.198$ & -- & -- & -- \\
        \rowcolor{myblue}
        Humanoid & Ours (opt) & $2.212\pm0.002$ & $2.293\pm0.003$ & $3.034\pm0.050$ & $13.129\pm0.123$ \\
        \addlinespace[3pt]
        \rowcolor{myblue}
        Grasp & Ours (LM-mesh) & $201.67\pm20.02$ & $275.31\pm8.70$ & $592.86\pm3.62$ & $4088.45\pm2.52$ \\
        \bottomrule

    }
    \par\vspace{4pt}
\end{table}

\nopagebreak[4]
\noindent
\textbf{Batch latency across other tasks.}
Table~\ref{tab:batch_scaling} shows efficient scaling for online retargeting: increasing the batch from $1$ to $1000$ raises RoboKit's hand latency from $1.37$ to $3.02$\,ms and humanoid latency from $2.21$ to $13.13$\,ms. RoboKit is faster than PyRoKi at every measured hand batch size, while Dex-Retargeting and GMR lack batched APIs. With four trajectory candidates per query for both methods, RoboKit is faster than cuRobo at every measured motion-planning batch size. Grasp optimization processes $1000$ initializations in $4.09$\,s.

Online timings are per frame across all streams; motion and grasp timings cover the full batch. These latency sweeps use different workloads from Tab.~\ref{tab:metrics_abcde}. cuRobo's graph search is disabled to retain four candidates. A dash denotes an unavailable batched API; n.m. denotes an unmeasured batch size.

\subsection{Comparison with Learned Policies}

Beyond the optimization baselines in the main paper, we compare our distilled policies against dedicated learned methods. For hand motion retargeting, we evaluate against GeoRT~\cite{yin2025geometric} on DexYCB data.

\begin{table}[H]
    \tablestyle
    \caption{\textbf{Hand retargeting} against GeoRT~\cite{yin2025geometric} on DexYCB.}
    \label{tab:geort}
    \metricstable{l ccc}{4}{%
        \toprule
        Method & \makecell{Tip\\(mm) $\downarrow$} & \makecell{PA-Tip\\(mm) $\downarrow$} & \makecell{P95\\(mm) $\downarrow$} \\
        \midrule
        GeoRT~\cite{yin2025geometric} & 37.4                  & 9.5                      & 58.5                  \\
        \rowcolor{myblue}
        Ours (distilled)              & \textbf{23.0}         & \textbf{7.8}             & \textbf{29.9}         \\
        \bottomrule

    }
    \par\vspace{4pt}
\end{table}

\nopagebreak[4]
\noindent
\textbf{Comparison with learned policies.}
We report fingertip position error (Tip), Procrustes-aligned tip error (PA-Tip), and the 95th-percentile tip error (P95), all in mm.

\noindent
Our distilled policy, trained with geometry-aware retargeting costs, outperforms GeoRT on all three metrics (Tab.~\ref{tab:geort}), reducing the fingertip position error from $37.4$ to $23.0$\,mm and the 95th-percentile error from $58.5$ to $29.9$\,mm. For grasp synthesis, our distilled policy already surpasses the learned baselines UGG~\cite{lu2024ugg} and Dex1B~\cite{ye2025dex1b} (see the main paper).

\subsection{Comparison with DIAL-MPC}
\label{sec:dial}

We compare against DIAL-MPC~\cite{xue2025dial}, a sampling-based MPC for torque-level legged locomotion, on the three Go2 tasks of its Table~I: velocity tracking with a trot gait, a sequence of $10$ random jumps, and climbing onto a crate $0.125$--$0.6$\,m tall ($10$ heights, two seeds each). Both methods use DIAL-MPC's released models, rewards and control settings. RoboKit uses one configuration for all three tasks, with $0.6\times$ DIAL-MPC's samples. We report the tracking error, DIAL-MPC's jump score, and crate success (all four feet on top, torso upright). Time is wall-clock per control step.

\begin{table}[H]
    \tablestyle
    \caption{\textbf{Locomotion MPC} against DIAL-MPC~\cite{xue2025dial} on Go2; time per control step.}
    \label{tab:dial}
    \metricstable{l cc cc cc}{7}{%
        \toprule
         & \multicolumn{2}{c}{Trot} & \multicolumn{2}{c}{Seq. Jump} & \multicolumn{2}{c}{Crate Climb} \\
        \cmidrule(lr){2-3} \cmidrule(lr){4-5} \cmidrule(lr){6-7}
        Method & \makecell{Track.\\err. $\downarrow$} & \makecell{Time\\(ms) $\downarrow$} & \makecell{Score\\$\uparrow$} & \makecell{Time\\(ms) $\downarrow$} & \makecell{Succ.\\(\%) $\uparrow$} & \makecell{Time\\(ms) $\downarrow$} \\
        \midrule
        DIAL-MPC~\cite{xue2025dial} & 0.0030 & 74.4 & 0.885 & 65.7 & 75 & 1354 \\
        \rowcolor{myblue}
        Ours & \textbf{0.0025} & \textbf{34.8} & \textbf{0.920} & \textbf{29.1} & \textbf{80} & \textbf{163} \\
        \bottomrule
    }
    \par\vspace{4pt}
\end{table}

\noindent
RoboKit matches or exceeds DIAL-MPC on all three tasks while running $2.1\times$, $2.3\times$ and $8.3\times$ faster (Tab.~\ref{tab:dial}).

\Needspace{16\baselineskip}
\subsection{Grasp Synthesis Ablation}

We compare mesh and SDF geometry representations for grasp synthesis on $4{,}096$ grasps ($128$ scenes $\times$ $32$ initializations).

\begin{table}[H]
    \tablestyle
    \caption{\textbf{Grasp synthesis ablation} on $4{,}096$ grasps (mean$\pm$population std across $5$ runs).}
    \label{tab:grasp_run_statistics}
    \metricstable{lcccc}{5}{%
        \toprule
        Method & \makecell{Time\\(s)} & \makecell{SR\\(\%)} & Q1 & \makecell{Pen.\\(mm)} \\
        \midrule
        LM-mesh & $16.605\pm0.029$ & $48.49\pm0.56$ & $0.0702\pm0.0004$ & $0.547\pm0.007$ \\
        GD-mesh & $18.412\pm0.378$ & $41.89\pm0.89$ & $0.0683\pm0.0013$ & $1.441\pm0.017$ \\
        LM-sdf & $2.973\pm0.033$ & $46.83\pm0.41$ & $0.0666\pm0.0007$ & $0.702\pm0.014$ \\
        GD-sdf & $4.334\pm0.007$ & $41.52\pm0.78$ & $0.0657\pm0.0012$ & $1.485\pm0.017$ \\
        \bottomrule

    }
    \par\vspace{4pt}
\end{table}

\nopagebreak[4]
\noindent
\textbf{Grasp synthesis ablation.}
Values are mean$\pm$population std across $5$ runs, using the original full evaluation. Time covers optimizer iterations; success requires pre-grasp penetration at most $5$\,mm and passing the ManiSkill grasp test under external forces. LM uses $64$ iterations and annealed GD uses $6{,}000$.

\Needspace{16\baselineskip}
\subsection{Comparison to CPU IK Baselines}

We compare our solver against Pinocchio~\cite{carpentier2019pinocchio} and Pink~\cite{pink}, two widely-used CPU-based Python IK libraries, using their official example scripts (Tab.~\ref{tab:single_instance_ik}).

\begin{table}[H]
    \tablestyle
    \caption{\textbf{Comparison to CPU IK baselines} on the 7-DoF Franka Panda arm.}
    \label{tab:single_instance_ik}
    \metricstable{l cccc}{5}{%
        \toprule
        Method & \makecell{Time\\(ms)} & \makecell{SR\\(\%)} & \makecell{Pos.\ Err\\(mm)} & \makecell{Ori.\ Err\\(rad)} \\
        \midrule
        Pinocchio~\cite{carpentier2019pinocchio} & 3.35          & 95.31        & 1.923                 & $4.53 \times 10^{-3}$ \\
        Pink~\cite{pink}                         & 10.80         & 79.69        & $3.95 \times 10^{-4}$ & $4.28 \times 10^{-7}$ \\
        \rowcolor{myblue}
        Ours (opt)                               & \textbf{2.36} & \textbf{100} & $4.05 \times 10^{-4}$ & $4.78 \times 10^{-7}$ \\
        \bottomrule

    }
    \par\vspace{4pt}
\end{table}

\nopagebreak[4]
\noindent
\textbf{Comparison to CPU IK baselines.}
Our solver leverages the GPU for random seed sampling and the Levenberg--Marquardt iterations. Both our method and Pink enforce joint-limit constraints, whereas the default Pinocchio IK example does not. Thanks to multi-seed sampling, our method achieves a $100\%$ success rate, compared to $95.31\%$ for Pinocchio and $79.69\%$ for Pink, while also being the fastest at $2.36$\,ms ($1.4\times$ faster than Pinocchio, $4.6\times$ faster than Pink). Solution accuracy is comparable to Pink and roughly an order of magnitude better than Pinocchio in both position and orientation error.

\section{Implementation Details}

\subsection{Analytic Objectives}

Table~\ref{tab:all_terms} summarizes the pre-built analytic objectives; we detail their mathematical formulations below.

\begin{table}[htbp]
    \tablestyle
    \caption{\textbf{Pre-built analytic objectives.}}
    \label{tab:all_terms}
    \setlength{\tabcolsep}{4pt}
    \begin{tabularx}{\linewidth}{l>{\raggedright\arraybackslash}Xc}
        \toprule
        \textbf{Term} & \textbf{Purpose} & \makecell{\textbf{Analytic}\\\textbf{Jacobian}} \\
        \midrule
        \multicolumn{3}{l}{\textit{Task Objectives}}                                                             \\
        Pose Matching       & Match link SE(3) pose to target via log map           & \checkmark                 \\
        Retargeting         & Pairwise link-to-keypoint alignment                   & \checkmark                 \\
        Contact Attraction  & Attract collision spheres to object contact points    & \checkmark                 \\
        Force Closure       & Penalize net contact wrench for grasp equilibrium     & \checkmark                 \\
        Mask Alignment      & Match rendered robot masks to observed silhouettes    & \checkmark$^*$             \\
        \midrule
        \multicolumn{3}{l}{\textit{Safety \& Collision}}                                                         \\
        Collision Avoidance & Penalize robot sphere penetration into mesh SDF       & \checkmark                 \\
        Self-Collision      & Penalize inter-link sphere--sphere penetration        & \checkmark                 \\
        \midrule
        \multicolumn{3}{l}{\textit{Joint \& Derivative Limits}}                                                  \\
        Joint Limit         & Penalize joint positions outside mechanical bounds    & \checkmark                 \\
        Velocity Limit      & Penalize joint velocities exceeding limits            & \checkmark                 \\
        Acceleration/Jerk Limits & Penalize trajectory joint derivatives outside bounds & \checkmark              \\
        \midrule
        \multicolumn{3}{l}{\textit{Regularization \& Smoothness}}                                                \\
        Smoothness          & Penalize joint/base changes between consecutive steps & \checkmark                 \\
        EE Pose Smoothness  & Penalize EE SE(3) pose changes across trajectory      & \checkmark                 \\
        Rest Pose           & Regularize joints toward a rest configuration         & \checkmark                 \\
        Start Config        & Penalize deviation from initial trajectory config     & \checkmark                 \\
        Base Damping        & Penalize floating-base motion from reference pose     & \checkmark                 \\
        \midrule
        \multicolumn{3}{l}{\textit{Performance}}                                                                 \\
        Manipulability      & Maximize Yoshikawa manipulability at target link      & \checkmark                 \\
        \bottomrule

    \end{tabularx}
    \par\vspace{4pt}
\end{table}

All terms support batched evaluation. Mask alignment ($*$) combines analytical projection derivatives with central-difference image gradients; the other listed terms use analytical Jacobians.

\noindent
\textbf{Mathematical Formulations}
The solver minimizes the sum of squared weighted residuals:
$$
    \min_{q} \;\frac{1}{2}\sum_{k} \bigl\| W_k \, c_k(q) \bigr\|_2^2,
$$
where $c_k(q)$ is an unweighted residual vector and $W_k = \mathrm{diag}(\alpha_k \, w_k)$ combines the task gain $\alpha_k \in (0,1]$ with per-dimension residual weights $w_k$. Task inputs are omitted here for brevity.

\noindent
\textbf{Pose Matching Cost.} The pose matching cost penalizes the SE(3) error between the current link pose and a target:
$$
    c_{\mathrm{pose}}(q,\, i,\, {}^{w}\mathbf{T}_{\mathrm{tgt}}) =
    \log\!\bigl({}^{w}\mathbf{T}_{i}^{-1}\, {}^{w}\mathbf{T}_{\mathrm{tgt}}\bigr) \in \mathbb{R}^6,
$$
where $\log : \mathrm{SE}(3) \to \mathfrak{se}(3)$ is the logarithm map. The first three components are weighted by $w_{\mathrm{pos}}$ and the last three by $w_{\mathrm{ori}}$. The analytical Jacobian is:
$$
    \frac{\partial c_{\mathrm{pose}}}{\partial q} =
    -\,J_{\log}\!\bigl({}^{w}\mathbf{T}_{\mathrm{tgt}}^{-1}\, {}^{w}\mathbf{T}_{i}\bigr)\,
    \mathrm{Ad}({}^{w}\mathbf{T}_{i}^{-1})\,
    \mathbf{S}_i,
$$
where $J_{\log}$ is the Jacobian of the SE(3) log map, $\mathrm{Ad}(\cdot)$ is the adjoint representation, and $\mathbf{S}_i$ is the world-frame geometric Jacobian of link $i$.

\noindent
\textbf{Retargeting Cost.} The retargeting cost measures pairwise correspondence between robot link positions $\{p^{\mathrm{r}}_k\}$ and target keypoints $\{p^{\mathrm{t}}_k\}$. For each pair $(i, j)$:
$$
    c_{\mathrm{retarget}}(q) = \bigl(p^{\mathrm{t}}_i - p^{\mathrm{t}}_j\bigr) - s \cdot \bigl(p^{\mathrm{r}}_i - p^{\mathrm{r}}_j\bigr),
$$
where $s$ is a position scale factor. An optional angular term penalizes direction mismatch of the displacement vectors.

\noindent
\textbf{Contact Attraction Cost.} The contact attraction cost encourages robot collision spheres to reach object contact points:
$$
    c_{\mathrm{contact}}(q,\, p_{\mathrm{obj}}) = \max\!\bigl(0,\; \min_{i}\bigl(\|p_{\mathrm{obj}} - {}^{w}c_i\| - r_i\bigr)\bigr),
$$
where ${}^{w}c_i$ and $r_i$ are the world-frame center and radius of the $i$-th collision sphere. The Jacobian propagates through only the closest active sphere via $-\hat{d}^\top \frac{\partial {}^{w}c_{i^*}}{\partial q}$, where $\hat{d}$ is the unit direction from sphere to contact point.

\noindent
\textbf{Force Closure Cost.} The force closure cost drives the net contact wrench toward zero, encouraging the grasp to resist arbitrary external loads:
$$
    c_{\mathrm{fc}}(q) = \begin{bmatrix} \sum_{k} \hat{n}_k \\[4pt] \sum_{k} p_k \times \hat{n}_k \end{bmatrix} \in \mathbb{R}^6,
$$
where $\hat{n}_k$ is the object-surface unit normal at the $k$-th contact (queried dynamically from the object geometry) and $p_k$ is the world-frame contact position. The first three components penalize net force imbalance and the last three penalize net torque imbalance. With $N_k = \partial\hat{n}_k/\partial p_k$, the analytical Jacobian is:
$$
    \frac{\partial c_{\mathrm{fc}}}{\partial q}
    = \begin{bmatrix}
        \sum_k N_k\frac{\partial p_k}{\partial q} \\[4pt]
        \sum_k \left(\frac{\partial p_k}{\partial q}\times\hat{n}_k
        + p_k\times N_k\frac{\partial p_k}{\partial q}\right)
      \end{bmatrix}.
$$
Voxel queries differentiate the normalized trilinear SDF gradient analytically, and primitive queries use analytical normal derivatives. Mesh queries return triangle face normals, so $N_k=0$ within the selected face. Derivatives are discontinuous at voxel-cell, face, and closest-object switches; no finite-difference normal queries are used.

\noindent
\textbf{Mask Alignment Cost.} For rendered and observed masks $M^{\mathrm{r}}$ and $M^{\mathrm{o}}$ at the current pyramid level, including its configured blur, the per-pixel residual is:
$$
    c_{\mathrm{mask},uv} = M^{\mathrm{r}}_{uv} - M^{\mathrm{o}}_{uv}.
$$
For a camera-frame twist $\boldsymbol{\xi}$, the image-motion Jacobian is:
$$
    \frac{\partial c_{\mathrm{mask},uv}}{\partial\boldsymbol{\xi}}
    = \begin{bmatrix} \partial_u M^{\mathrm{r}} & \partial_v M^{\mathrm{r}} \end{bmatrix}
      \frac{\partial(u,v)}{\partial\boldsymbol{\xi}}.
$$
The projection derivative is analytical; image gradients use central pixel differences, e.g., $\partial_u M^{\mathrm{r}}_{uv}=(M^{\mathrm{r}}_{u+1,v}-M^{\mathrm{r}}_{u-1,v})/2$ for interior pixels. An SE(3) adjoint maps the camera-frame Jacobian to the calibration variable's tangent frame. Per-observation weights scale the squared residuals, and the implementation accumulates $J^\top J$ and $J^\top c$ over pixels without storing the full Jacobian. Optional valid depth observations add a Huber residual.

\noindent
\textbf{Collision Avoidance Cost.} The robot is approximated by collision spheres $s_i = (c_i, r_i)$, and each object $\mathcal{O}$ by the hybrid voxel-SDF and mesh representation (Sec.~\ref{subsec:accel}), which gates an exact mesh query to a near-surface band and otherwise reads a cheap baked voxel SDF:
$$
    c_{\mathrm{coll}}(q) = \max\!\bigl(0,\; r_i + \epsilon - \mathrm{SDF}({}^{w}c_i,\, \mathcal{O})\bigr),
$$
where $\mathrm{SDF}(\cdot, \mathcal{O})$ returns the signed distance to object $\mathcal{O}$ (negative inside), and $\epsilon \ge 0$ is a safety margin. When multiple objects are present, $\mathrm{SDF}_{\mathrm{eff}} = \min_{\mathcal{O}} \mathrm{SDF}(\cdot, \mathcal{O})$. The Jacobian is $-\hat{n}_i^\top \frac{\partial {}^{w}c_i}{\partial q}$ when the constraint is active, where $\hat{n}_i$ is the outward surface normal and $\frac{\partial {}^{w}c_i}{\partial q} = \mathbf{v} + \boldsymbol{\omega} \times {}^{w}c_i$ from the spatial motion subspace.

\noindent
\textbf{Self-Collision Cost.} For non-adjacent link pairs $(i, j)$, the self-collision cost penalizes sphere overlap:
$$
    c_{\mathrm{self}}(q) = \max\!\bigl(0,\; r_i + r_j + \epsilon - \|{}^{w}c_i - {}^{w}c_j\|\bigr).
$$
The Jacobian uses the unit direction between sphere centers: $-\hat{d}_{ij}^\top(\frac{\partial {}^{w}c_i}{\partial q} - \frac{\partial {}^{w}c_j}{\partial q})$.

\noindent
\textbf{Joint Limit Cost.} The joint limit cost penalizes configurations exceeding mechanical bounds:
$$
    c_{\mathrm{limit}}(q) = \max(0,\, q - q_{\mathrm{upper}}) + \max(0,\, q_{\mathrm{lower}} - q).
$$
The Jacobian is a diagonal matrix of signs: $+1$ above the upper limit, $-1$ below the lower limit, and $0$ otherwise. Hard QP constraints are also available: $q_{\mathrm{lower}} - q \le \Delta q \le q_{\mathrm{upper}} - q$.

\noindent
\textbf{Velocity Limit Cost.} The velocity limit cost penalizes excessive joint velocities:
$$
    c_{\mathrm{vel}}(q_t, q_{t-1}) = \max\!\left(0,\; \left|\frac{q_t - q_{t-1}}{\Delta t}\right| - \dot{q}_{\max}\right).
$$
The Jacobian is $\frac{\mathrm{sign}(\dot{q})}{\Delta t}$ when the limit is violated.

\noindent
\textbf{Acceleration and Jerk Limit Costs.} For derivative order $p\in\{2,3\}$, trajectory limits penalize acceleration or jerk outside per-joint bounds $[l_j^{(p)},u_j^{(p)}]$:
$$
    d_{t,j}^{(p)} = \frac{1}{\Delta t^p}\sum_{k=-2}^{2}\beta_k^{(p)}q_{t+k,j},
    \qquad
    c_{t,j}^{(p)} = \max\!\left(0,\;l_j^{(p)}-d_{t,j}^{(p)},\;d_{t,j}^{(p)}-u_j^{(p)}\right).
$$
The default five-point central stencils, ordered from $k=-2$ to $2$, are:
$$
    \boldsymbol{\beta}^{(2)} = \tfrac{1}{12}(-1,16,-30,16,-1),
    \qquad
    \boldsymbol{\beta}^{(3)} = (-\tfrac12,1,0,-1,\tfrac12).
$$
Boundary indices are clamped to the first or last frame, holding endpoint values. Each sparse Jacobian entry is $\sigma_{t,j}^{(p)}\beta_k^{(p)}/\Delta t^p$, where $\sigma=-1$ below the lower bound, $+1$ above the upper bound, and $0$ otherwise; entries mapping to the same held frame add together. Forward-difference stencils are also supported. Revolute-joint acceleration and jerk bounds have units rad/s$^2$ and rad/s$^3$, respectively.

\noindent
\textbf{Smoothness Cost.} The smoothness cost penalizes large joint changes between consecutive time steps:
$$
    c_{\mathrm{smooth}}(q_t, q_{t-1}) = q_t - q_{t-1}.
$$
For floating-base robots, an additional base term is $\log({}^{w}\mathbf{T}_{b,t}\, {}^{w}\mathbf{T}_{b,t-1}^{-1})$. The joint Jacobian is the identity matrix; the base Jacobian is $J_{\log}(T_{\mathrm{error}}) \cdot \mathrm{Ad}({}^{w}\mathbf{T}_{b,t})$.

\noindent
\textbf{EE Pose Smoothness Cost.} Penalizes changes in end-effector SE(3) poses between consecutive trajectory frames:
$$
    c_{\mathrm{ee}}(t, k) = \log\!\bigl({}^{w}\mathbf{T}_{k,t}^{-1}\, {}^{w}\mathbf{T}_{k,t-1}\bigr) \in \mathbb{R}^6,
$$
for each end-effector $k$ and consecutive frames $t$, $t{-}1$.

\noindent
\textbf{Rest Pose Cost.} The rest pose cost regularizes joints toward a preferred configuration:
$$
    c_{\mathrm{rest}}(q) = q - q_{\mathrm{rest}}.
$$
For floating-base robots, an additional term is $\log({}^{w}\mathbf{T}_b\, {}^{w}\mathbf{T}_{\mathrm{rest}}^{-1})$. The joint Jacobian is the identity matrix.

\noindent
\textbf{Start Configuration Cost.} For trajectory optimization, penalizes deviation of the first frame from a specified initial configuration: $c_{\mathrm{start}}(q_0) = q_0 - q_{\mathrm{init}}$.

\noindent
\textbf{Base Damping Cost.} Discourages floating-base motion from a reference pose:
$$
    c_{\mathrm{damp}}({}^{w}\mathbf{T}_b) = \log\!\bigl({}^{w}\mathbf{T}_b\, {}^{w}\mathbf{T}_{\mathrm{ref}}^{-1}\bigr) \in \mathbb{R}^6.
$$

\noindent
\textbf{Manipulability Cost.} Maximizes the Yoshikawa measure at a target link:
$$
    m(q) = \sqrt{\det\!\bigl(\mathbf{J}_p\, \mathbf{J}_p^\top\bigr)}, \qquad
    c_{\mathrm{manip}}(q) = \frac{1}{m(q) + \varepsilon},
$$
where $\mathbf{J}_p \in \mathbb{R}^{3 \times n}$ is the translational Jacobian at the target link. The analytical gradient uses the cofactor matrix of $\mathbf{J}_p \mathbf{J}_p^\top$ and the derivative of $\mathbf{J}_p$ through the kinematic chain via spatial adjoints.

\newpage
\subsection{Algorithm Summary}
\label{subsec:algorithm}

We summarize the full RoboKit pipeline in three stages: parallel kinematic optimization (Algorithm~\ref{alg:solver}), offline policy distillation from filtered demonstrations (Algorithm~\ref{alg:offline}), and online policy fine-tuning through differentiable objectives (Algorithm~\ref{alg:online}). Algorithm~\ref{alg:hybrid} combines these stages sequentially: offline supervised training on filtered demonstrations initializes the policy, followed by fine-tuning with the online objective on freshly sampled task instances.

\begin{algorithm}[!ht]
    \caption{Parallel Geometry-Aware Kinematic Solving}
    \label{alg:solver}
    \begin{algorithmic}[1]
        \REQUIRE Residuals and weights $\{(c_k, W_k)\}$, task distribution $p(o)$, number of particles $M$, LM iterations $T_{\mathrm{opt}}$, number of rounds $R$, top-$K$ selection ratio
        \ENSURE Solution dataset $\hat{\mathcal{D}} = \{(o_i, q_i^*)\}$

        \FOR{each task instance $o_i \sim p(o)$}
        \STATE Sample $M$ initial particles $\{q^{(m)}\}_{m=1}^{M}$
        \FOR{$r = 1$ \TO $R$}
        \FOR{$t = 1$ \TO $T_{\mathrm{opt}}$}
        \STATE Evaluate weighted residuals $r_k^{(m)} = W_k \, c_k(q^{(m)}; o_i)$ for all $k, m$
        \STATE Compute analytical sparse Jacobians $J_k^{(m)}$
        \STATE Update each particle via Levenberg--Marquardt step
        \ENDFOR
        \STATE Rank particles by cost: $\mathcal{S}^{(m)} = \frac{1}{2}\sum_k \|r_k^{(m)}\|_2^2$
        \STATE Select top-$K$ particles $\{q^{(m)}\}_{m=1}^{K}$
        \IF{$r < R$ \textbf{and} resampling enabled}
        \STATE Resample $M$ particles from top-$K$
        \ENDIF
        \ENDFOR
        \STATE Select best solution: $q_i^* = \arg\min_m \mathcal{S}^{(m)}$
        \STATE Store $(o_i, q_i^*)$ in $\hat{\mathcal{D}}$
        \ENDFOR
        \RETURN $\hat{\mathcal{D}}$
    \end{algorithmic}
\end{algorithm}

\begin{algorithm}[!ht]
    \caption{Offline Policy Distillation}
    \label{alg:offline}
    \begin{algorithmic}[1]
        \REQUIRE Candidate dataset $\hat{\mathcal{D}}$ from Algorithm~\ref{alg:solver}, filtering criterion $\mathcal{F}$, policy $\pi_\theta$, learning rate $\eta$, training iterations $T_{\mathrm{off}}$
        \ENSURE Policy $\pi_\theta$

        \STATE Filter: $\mathcal{D} = \{(o_i, q_i^*) \in \hat{\mathcal{D}} \mid \mathcal{F}(o_i, q_i^*) = \text{True}\}$
        \FOR{$t = 1$ \TO $T_{\mathrm{off}}$}
        \STATE Sample mini-batch $(o, q^*) \sim \mathcal{D}$
        \STATE Compute $\mathcal{L}_{\mathrm{offline}}(\theta) = \|\pi_\theta(o) - q^*\|^2$
        \STATE Update $\theta \leftarrow \theta - \eta \, \nabla_\theta \mathcal{L}_{\mathrm{offline}}(\theta)$
        \ENDFOR
        \RETURN $\pi_\theta$
    \end{algorithmic}
\end{algorithm}

\begin{algorithm}[!ht]
    \caption{Online Policy Fine-Tuning}
    \label{alg:online}
    \begin{algorithmic}[1]
        \REQUIRE Residuals and weights $\{(c_k, W_k)\}$, task distribution $p(o)$, pretrained policy $\pi_\theta$ from Algorithm~\ref{alg:offline}, learning rate $\eta$, training iterations $T_{\mathrm{on}}$
        \ENSURE Policy $\pi_\theta$

        \FOR{$t = 1$ \TO $T_{\mathrm{on}}$}
        \STATE Sample task instances $o \sim p(o)$
        \STATE Predict $\hat{q} = \pi_\theta(o)$
        \STATE Compute objective loss $\mathcal{L}(\theta) = \frac{1}{2}\sum_k \|W_k \, c_k(\hat{q}; o)\|_2^2$
        \STATE Update $\theta \leftarrow \theta - \eta \, \nabla_\theta \mathcal{L}(\theta)$
        \ENDFOR
        \RETURN $\pi_\theta$
    \end{algorithmic}
\end{algorithm}

\begin{algorithm}[!ht]
    \caption{Hybrid Policy Distillation (Offline Warm-Up, Online Fine-Tuning)}
    \label{alg:hybrid}
    \begin{algorithmic}[1]
        \REQUIRE Candidate dataset $\hat{\mathcal{D}}$ from Algorithm~\ref{alg:solver}, filtering criterion $\mathcal{F}$, residuals and weights $\{(c_k,W_k)\}$, task distribution $p(o)$, policy $\pi_\theta$, learning rate $\eta$, offline iterations $T_{\mathrm{off}}$, online iterations $T_{\mathrm{on}}$
        \ENSURE Policy $\pi_\theta$

        \STATE Train $\pi_\theta$ on filtered demonstrations using Algorithm~\ref{alg:offline} for $T_{\mathrm{off}}$ iterations
        \STATE Fine-tune the resulting $\pi_\theta$ with the online objective using Algorithm~\ref{alg:online} for $T_{\mathrm{on}}$ iterations
        \RETURN $\pi_\theta$
    \end{algorithmic}
\end{algorithm}
\hfill

\clearpage
\section{Additional Visualization}
\label{sec:additional_viz}

\begin{figure*}[htbp]
    \centering
    \includegraphics[width=0.9\linewidth]{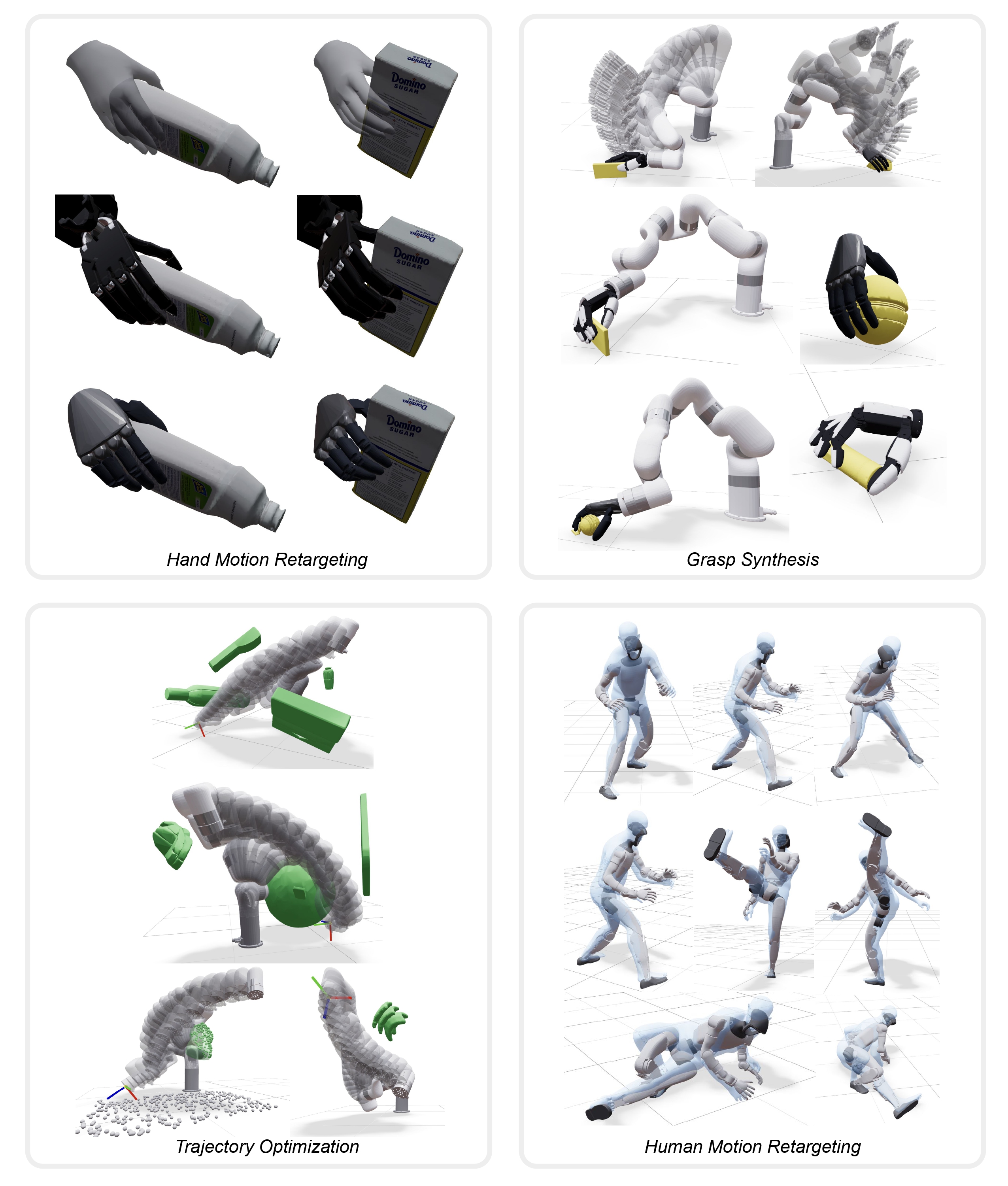}
    \caption{
        Qualitative results of RoboKit across core tasks.
        A single composite figure showcasing hand motion retargeting, grasp synthesis, trajectory optimization, and human motion retargeting.
    }
    \label{fig:add_visuals}
\end{figure*}

We provide additional visualization in Fig.~\ref{fig:add_visuals}.

\end{document}